\pdfoutput=1

\documentclass[11pt]{article}

\usepackage[preprint]{acl}

\usepackage{times}
\usepackage{latexsym}

\usepackage[T1]{fontenc}
\usepackage{kotex}

\usepackage[utf8]{inputenc}

\usepackage{microtype}

\usepackage{inconsolata}

\usepackage{graphicx}
\usepackage{xcolor}

\usepackage{booktabs}

\title{Evaluating Legal Reasoning Traces with Legal Issue Tree Rubrics}

\author{Jinu Lee\textsuperscript{1}, Kyoung-Woon On\textsuperscript{2}, Simeng Han\textsuperscript{3}, Arman Cohan\textsuperscript{4}, Julia Hockenmaier\textsuperscript{1} \\
  \textsuperscript{1}University of Illinois Urbana-Champaign \ \textsuperscript{2}LBOX \\
  \textsuperscript{3}Stanford \ \textsuperscript{4}Yale \\
  \texttt{\{jinulee2, juliahmr\}@illinois.edu}\ \ \texttt{kyoungwoon.on@lbox.kr} \ \ \texttt{\{shan6\}@stanford.edu}}

\begin{document}

\maketitle
\begin{abstract}
Evaluating the quality of LLM-generated reasoning traces in expert domains (\textit{e.g.}, law) is essential for ensuring credibility and explainability, yet remains challenging due to the inherent complexity of such reasoning tasks. We introduce LEGIT (LEGal Issue Trees), a novel large-scale (24K instances) expert-level legal reasoning dataset with an emphasis on reasoning trace evaluation. We convert court judgments into hierarchical trees of opposing parties' arguments and the court's conclusions, which serve as rubrics for evaluating the issue coverage and correctness of the reasoning traces. We verify the reliability of these rubrics via human expert annotations and comparison with coarse, less informative rubrics. Using the LEGIT dataset, we show that (1) LLMs' legal reasoning ability is seriously affected by both legal issue coverage and correctness, and that (2) retrieval-augmented generation (RAG) and RL with rubrics bring complementary benefits for legal reasoning abilities, where RAG improves overall reasoning capability, whereas RL improves correctness albeit with reduced coverage.\footnote{We fully release the \href{https://github.com/jinulee-v/LEGIT}{code} and \href{https://huggingface.co/datasets/jinulee-v/legit_ko_verl}{data}.}
\end{abstract}

\section{Introduction}

Large language models (LLMs) can solve complex reasoning problems by generating an intermediate reasoning trace ("\textit{Chain-of-thoughts}") before outputting the final answer \cite{wei_chain--thought_2022, guha2025openthoughtsdatarecipesreasoning}. Evaluating the quality of these reasoning traces is crucial for understanding as well as improving the reasoning ability of LLMs, for example, through selecting the best responses or reinforcement learning \citep{lanham_measuring_2023, yao_tree_2023, han-etal-2024-p,  lee_evaluating_2025}. 
However, evaluating reasoning traces for expert-level tasks, \textit{e.g.}, law and medicine, requires substantial domain expertise that even advanced LLMs do not yet fully possess \citep{mishra_investigating_2025, wang_process-supervised_2025}. Moreover, automatic evaluators often lack an understanding of domain-specific and nuanced \textit{desirability} shaping human judgments \citep{kim_biggen_2025, starace2025paperbenchevaluatingaisability}. For instance, when asked to predict the outcome of a court judgment, a good reasoning trace should not only avoid logical and factual errors but also exhaustively identify and analyze legal issues \citep{yu_structured_2025}. This underscores the need for a more sophisticated evaluation strategy for expert-level, domain-specific reasoning traces.
We choose law as our testbed, as its textual and logical nature enables objective evaluation of reasoning traces compared to domains that rely on probabilistic and data-driven reasoning, \textit{e.g.}, medicine and finance.


\begin{figure*}
    \centering
    \includegraphics[width=\linewidth]{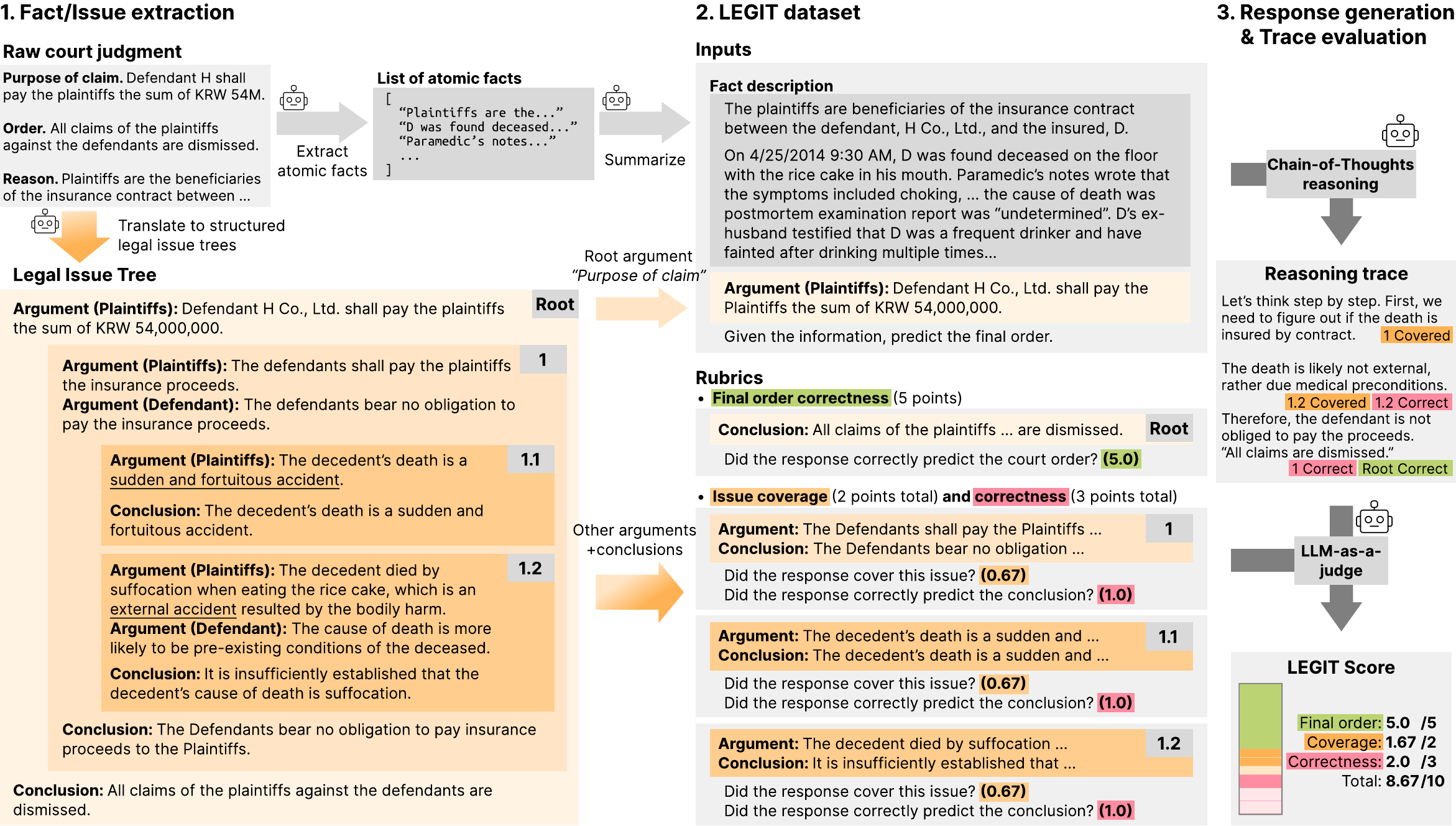}
    \caption{Overview of the LEGIT dataset and task. Facts and issue trees are extracted from real-world court judgments to serve as inputs and rubrics for the LEGIT task. See Appendix \ref{sec:appendix-example} for another example.}
    \label{fig:legit_overview}
\end{figure*}

In this paper, we introduce \textbf{LEGIT} (\textbf{LEGal Issue Trees}, Figure \ref{fig:legit_overview}), a large-scale Korean \textbf{legal judgment prediction (LJP)} benchmark with an additional emphasis on \textit{reasoning trace evaluation}. In contrast to other LJP datasets where the task is simply to predict the court's final order given basic facts \citep{cui_survey_2022}, LEGIT  includes additional \textbf{rubrics for reasoning traces} that are 
extracted from court judgments. These rubrics can be used to assess whether the reasoning traces fully cover the issues addressed during the trial (\textbf{issue coverage}) and if the reasoning about those issues is correct (\textbf{issue correctness}). They enable reliable LLM-as-a-judge evaluations of legal reasoning traces that are comparable to those of licensed lawyers, and facilitate principled and detailed analyses of the legal reasoning ability of LLMs.

Using LEGIT, we evaluate the legal reasoning ability of different LLMs, and discover that even state-of-the-art LLMs are not fully capable of complex legal judgment prediction. We identify and analyze two major error types: either failing to identify relevant issues (\textit{decomposition error}) or failing to reason correctly about facts (\textit{deduction error}).

Finally, we explore how retrieval augmented generation (RAG) and reinforcement learning (RL) affect legal reasoning performance. These two approaches show complementary benefits: RAG improves both issue coverage and correctness, while RL maximizes correctness at the expense of issue coverage.

\section{Background}

\subsection{Evaluating reasoning traces}

While reasoning traces dramatically increase LLM performance on complex tasks \citep{wei_chain--thought_2022}, they might contain factual or logical errors even if the final answer is correct \citep{lanham_measuring_2023}. Such errors can be critical in high-stakes fields such as law and medicine, if end-users rely on LLM responses for important decision-making \citep{mishra_investigating_2025}. Evaluating reasoning traces can help reduce this risk by filtering out incorrect responses or training the model via reinforcement learning \citep{lee_evaluating_2025}.

LLMs can be prompted to evaluate reasoning traces by finding errors or assessing the overall quality (\textit{"LLM-as-a-judge"}) \citep{yao_tree_2023}. Despite its simplicity, LLM-as-a-judge achieves strong performance in error detection and best response selection in math and programming \citep{huang_large_2024, kim_scaling_2025}. However, using generic evaluation criteria might not be sufficient to capture deeper insights into what constitutes a good reasoning trace. 
Defining desirability is particularly challenging in expert domains, where one must evaluate both precise factual knowledge and deeper domain-specific insights.

To incorporate diverse, fine-grained forms of desirability in reasoning traces, recent works introduced \textbf{instance-specific rubrics} \citep{gunjal_rubrics_2025, pathak_rubric_2025}. For instance, BigGen-Bench \citep{kim_biggen_2025} includes handcrafted rubrics tailored for individual problems that evaluate knowledge and problem-solving strategies. While such rubrics allow LLM judges to assess diverse criteria beyond correctness with high human-LLM agreement, the human effort required to define and validate rubrics hinders scaling this approach beyond evaluation-only benchmarks. By contrast, LEGIT automatically extracts high-quality rubrics from court judgments, resulting in scalable and expert-aligned evaluation of reasoning traces. 

\subsection{Legal Judgment Prediction (LJP)}

Legal judgment prediction (LJP) is the task of predicting the court's final order given the facts and claims of a case \citep{cui_survey_2022}. It is a representative legal reasoning task that simultaneously assesses an LLM's legal knowledge (statutes, case laws) and reasoning ability (applying laws to given facts)  \citep{jiang_legal_2023, huang_cmdl_2024}.

While traditional reasoning tasks like math often use final answer correctness as the sole criterion \citep{lightman_lets_2023, gao_llm_2025}, issue coverage and correctness are also critical in LJP \citep{yu_structured_2025}. Imagine a case where the beneficiary of an insurance policy (plaintiff) sues the insurance company (defendant) for payment, and the ground-truth order is to dismiss the case (Figure \ref{fig:legit_overview}). Even if two reasoning traces both predict that the case should be dismissed, if one thoroughly considers the contractual conditions of an insured accident (\textit{e.g.}, whether the accident was external) when the other does not, legal experts will find the former more useful in practice. Unfortunately, existing benchmarks account only for the final order prediction accuracy, and ignore 
both issue coverage and correctness \citep{aletras_predicting_2016, hwang_multi-task_2022, huang_cmdl_2024, adarsh_automating_2024}.

Current LJP benchmarks also fail to cover the full range of legal cases. As LJP was traditionally considered a classification/regression task, many works have focused on either criminal cases with a continuous spectrum of sentencing labels \citep{hwang_multi-task_2022, huang_cmdl_2024}, or a minimal set of cases with clear binary labels \citep{aletras_predicting_2016, adarsh_automating_2024}. To the best of our knowledge, no LJP dataset has covered arbitrary case types observed in civil and administrative cases, which account for up to 84.1\% of total court cases in the US \citep{united_state_court_federal_2024} and 69.4\% in Korea \citep{court_of_korea_2024_2024}.

\section{The LEGIT Dataset}
\label{sec:dataset}

\subsection{Legal issue trees}
Legal issues are inherently hierarchical \citep{jiang_legal_2023}. 
Returning to our example of beneficiaries suing an insurance company (Figure \ref{fig:legit_overview}), the \textit{purpose of claim}\footnote{The topmost claim, which only contains the subject-matter (money, property, disposition) and the action (payment, register, revocation) but not any of the underlying reason.} is that the company should pay the plaintiffs. The reason is that the event of death is insured by contract. To prove that the event is insured, the plaintiff must show that it is a sudden, fortuitous, and external accident. For instance, an event is external if the death is directly caused by an external event (\textit{e.g.}, choking), and not the consequence of pre-existing medical conditions.

This motivates structuring legal issues as a tree, where each node should include the arguments made by the parties and the judge’s conclusion regarding whether to accept or decline the argument. The tree structure reflects how a legal issue can be decomposed into child issues. While higher-level issues are bound to \textit{laws} like "If the accident is sudden, fortuitous, and external, the insurance company bears obligation to pay the beneficiary", lower issues require common sense inference on \textit{facts}, \textit{e.g.}, "If the deceased fainted after drinking multiple times before, it is likely that such pre-existing condition is the cause of death". The root node's argument is always the purpose of claim, and the conclusion corresponds to the court’s final order.  

We can therefore view the legal judgment prediction task as a \textit{backward chaining} (top-down traversal) of the legal issue tree, iterating between two procedures: identifying the relevant child issues from the given issue (\textit{decomposition}), and reasoning about an issue using the base facts and the conclusions of child issues (\textit{deduction}) \citep{kazemi_lambada_2023, lee_symba_2025}. 
This allows us to evaluate LLMs not just on the correctness of their final order prediction, but also on whether they can properly decompose issues into their children (\textit{issue coverage}) and can reason about them correctly (\textit{issue correctness}).


\subsection{Data source and statistics}
To construct LEGIT, we sample judgments issued by Korean District courts from the LBOX database (\url{lbox.kr}). We only gather judgments where the final order is \textit{deterministic} by law (Appendix~\ref{sec:appendix-dataset-preproc}), avoiding the cases where the final decisions are at the discretion of judges. For instance, in the Korean Criminal Act, the judge has the discretion to determine the sentence within a certain range (\textit{e.g.},  between 6 months and 2 years in jail). 
After filtering judgments that contain keywords related to such non-deterministic orders, the final dataset includes 24,406 judgments from diverse areas of civil and administrative law, which are divided into a training (24,106) and test (300) split. See Appendix~\ref{sec:appendix-dataset-stats} for more statistics.

\subsection{Fact extraction}
\label{sec:fact-extraction}
Facts serve as the core inputs for LLMs in the LJP task \citep{cui_survey_2022}. While existing legal datasets commonly use the "Facts" section of the judgment without further modification \citep{hwang_multi-task_2022, huang_cmdl_2024}, this section often neglects \textit{indirect facts} (which imply important facts by Rule of Thumb) and \textit{supplementary facts} (which (dis)prove the probative power of an evidence) that are essential for a complete investigation of a case. Therefore, instead of only using the "Facts" section, we use an LLM (Gemini-2.0-Flash) to (1) extract unit factual events in simple sentences from the entire judgment \citep{min_factscore_2023} and (2) generate a coherent case description from the list of unit facts, both with 1-shot examples (Appendix~\ref{sec:appendix-prompts}).

\subsection{Issue structure extraction}
\label{sec:issue-extraction}
We prompt LLMs to generate rich legal issue trees directly from the court judgment, using Gemini-2.0-Flash and 3-shot examples manually curated by the authors. To improve the quality and minimize errors, we run the LLM twice: first, to create the issue tree structure from raw judgments, and second, to refine the results with another prompt to eliminate errors and mistakes that were often observed in the first round (Appendix~\ref{sec:appendix-prompts}). 

As cases with more issues tend to involve more facts and relevant statutes, we divide the dataset into three subsets based on the number of issues (easy: $\leq$25\%, medium: 25-75\%, hard: $\geq$75\%). LEGIT's test split includes 300 examples, with 100 from each difficulty subset, where the paper authors have manually inspected and fixed errors in the dataset.

See Appendix~\ref{sec:appendix-dataset-preproc} for manual evaluation results of the automatic data construction pipeline.


\subsection{Issue-to-rubric conversion}
For LLM-as-a-judge evaluation, we convert issues into rubrics. The LLM judge will evaluate each issue individually. Specifically, for each extracted issue, the LLM jointly predicts if that issue was mentioned in the given response (coverage) and the response includes the correct conclusion of the given issue (correctness) with a short Chain-of-thoughts rationale. The scores are calculated by the point scheme described below.

\subsection{LEGIT score calculation}
The LEGIT score has three components: final order correctness, issue coverage, and issue correctness. \textbf{Final order correctness} represents whether the LLM-predicted final order is correct, which is the goal of the LJP task. Since LEGIT intentionally uses only deterministic judgments, the score is binary: $5$ points if the final order exactly matches the ground truth, and $0$ otherwise. \textbf{Issue coverage} (max: $2$ points) measures whether the response has identified the legal issues mentioned in the court judgment. If the issue tree has $N \geq 1$ non-root nodes, we assign $2/N$ points for each issue that is covered. Finally, for \textbf{issue correctness} (max: $3$ points), we assign $3/N$ points for each issue that is covered \textit{and} the predicted conclusion is correct. The maximum total LEGIT score is 10 points.

The score distribution between the three components is based on two core principles. First, final order correctness (5 points) is the most important criterion, as LJP's fundamental goal is to accurately predict the final order. Second, issue correctness (3/10) is more rewarded than issue coverage (2/10) because predicting the conclusion of the issue requires finding the exact relevant rationales and reasoning about them, but predicting the existence of an issue does not need to be as precise.

\subsection{Pointers to relevant appendices}

We present LEGIT data statistics and manual quality inspection results in Appendix \ref{sec:appendix-dataset}. Prompts used for data construction are included in Appendix \ref{sec:appendix-prompts}.

\section{Reliability of LLM-as-a-judge}
\label{sec:reliability}

In this section, we show that LEGIT rubrics allow reliable evaluation of reasoning traces by measuring inter-rater agreement of human experts and LLMs, and comparing with coarser rubrics.


\subsection{Inter-rater agreement}

\subsubsection{Experimental settings}

\paragraph{Reasoning trace generation} To begin with, we sample solutions from 12 frontier LLMs, including OpenAI \{o3, GPT-4.1, GPT-4.1-mini\} \citep{openai_gpt-4_2024}, Google Gemini \{2.5-Pro, 2.5-Flash, 2.0-Flash\} \citep{team_gemini_2025}, Gemma 3 \{27B, 12B, 4B\} \citep{team_gemma_2025}, EXAONE \{3.5-32B, 3.5-7.8B, 3.0-7.8B\} \citep{research_exaone_2024}\footnote{All models are instruction-tuned versions unless noted otherwise.}. A few widely used open-weight models, including Qwen, LLaMA, and DeepSeek, were discarded due to their inability to generate fluent Korean responses for LEGIT inputs. Each model was given the facts and the purpose of claim, and instructed to predict the final court order with emphasis on identifying possible issues in its reasoning trace. The temperature was set to 0 for reproducibility.

\paragraph{LLM-as-a-judge evaluation} We employ 10 different LLMs to evaluate reasoning traces with LEGIT rubrics. Specifically, we use OpenAI \{GPT-4.1, GPT-4.1-mini\}, Gemini \{2.5-Pro, 2.5-Flash, 2.0-Flash\}, Gemma 3 \{27B, 12B, 4B\}, and EXAONE \{3.5-7.8B, 3.5-2.4B\}.
For each issue, the LLM evaluators are prompted to assess if the given solution mentions the issue (issue coverage) and if the issue's conclusion is accurately predicted (issue correctness). We also set the temperature to 0 for evaluators.

\begin{figure}[t]
    \centering
    \includegraphics[width=\linewidth]{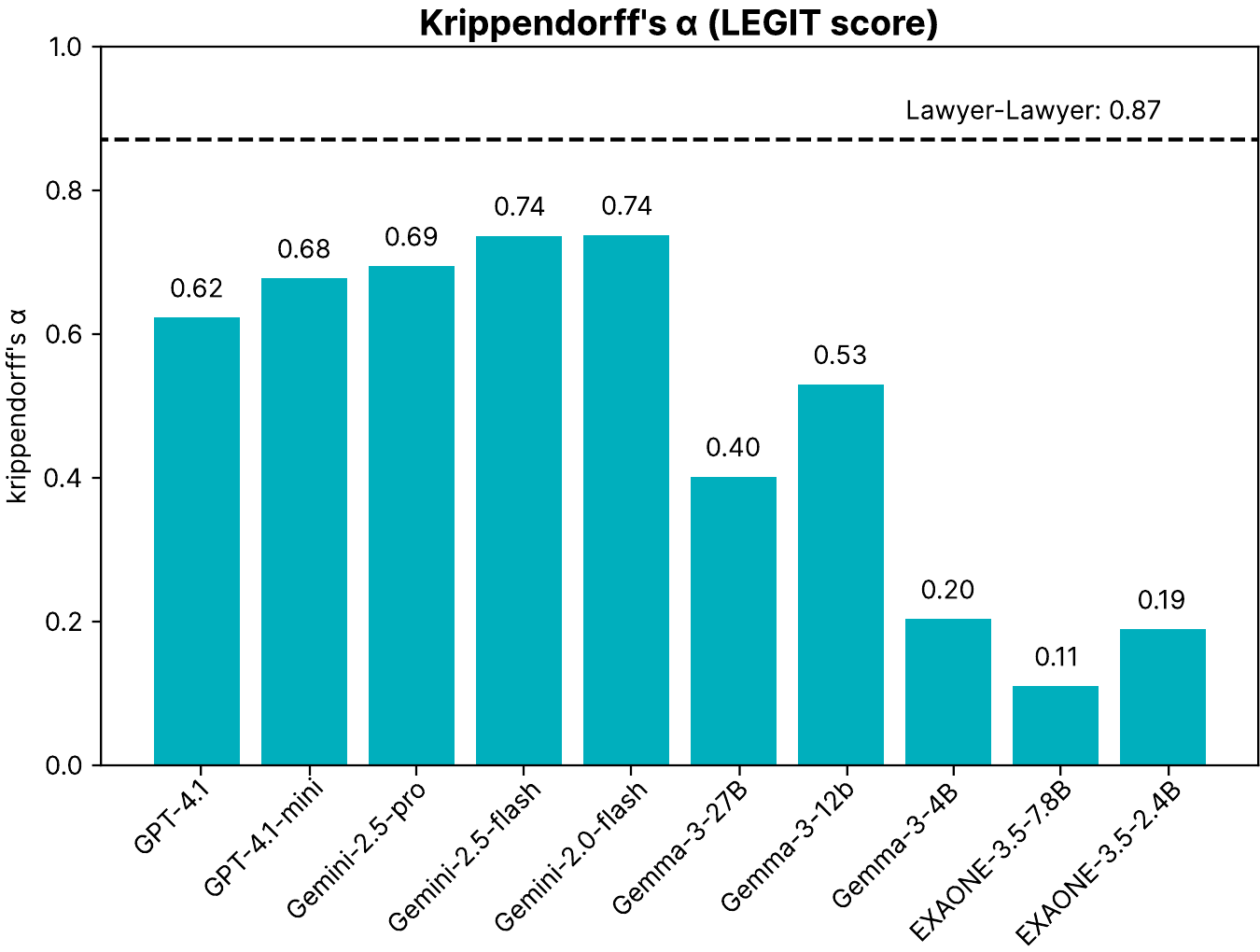}
    \caption{Lawyer-LLM inter-rater agreement in LEGIT score evaluation. Lawyers achieve strong agreement, ensuring that the generated rubrics are sound and effective. While strong LLMs (Gemini, GPT) achieve significant agreement with human experts, weaker open-sourced LLMs exhibit limited agreement.}
    \label{fig:llm_lawyer_agreement}
\end{figure}

\paragraph{Human expert evaluation} Two independent groups of licensed Korean lawyers were provided with 44 problems randomly sampled from the test split (/300), with responses obtained from a random LLM. The lawyers were instructed to annotate whether the LLM responses cover the given issue and reach the correct conclusion for the issue, mirroring the LLM-as-a-judge setting.

We use Krippendorff's $\alpha$ to compute the inter-rater agreement between lawyers and LLM evaluators, with the error function $\delta(s_1, s_2) = (s_1-s_2)^2$.

\subsubsection{Results}

\paragraph{Human experts achieve excellent inter-rater agreement with LEGIT rubrics.} First, we analyze how two lawyers agree on evaluating the LEGIT dataset. Lawyers achieve Krippendorff's $\alpha$ score of 0.87, where the typically recommended threshold is $\alpha$>0.67 \citep{krippendorff_reliability_2006, stefanovitch_holistic_2023}. Such a strong agreement indicates that the rubrics are objective and unambiguous to human experts, proving the clarity and reliability of LEGIT rubrics.

\paragraph{Strong LLMs can reliably evaluate reasoning traces with LEGIT rubrics.} The LLM-Lawyer agreement on LEGIT scores shows the reliability of LLM-as-a-judge evaluations. Closed-source models (GPT and Gemini) achieve Krippendorff's $\alpha$=0.62-0.74 with human experts. While larger open-weight models show reasonable agreement with lawyers, smaller open-weight models tend to demonstrate lower performance. For instance, Gemma-3-12B achieves Krippendorff's $\alpha$=0.53 while the 4B version only reaches $\alpha$=0.20, indicating that evaluating legal reasoning traces requires substantial domain knowledge and reasoning ability even with highly informative LEGIT rubrics.

\begin{figure}[t]
    \centering
    \includegraphics[width=\linewidth]{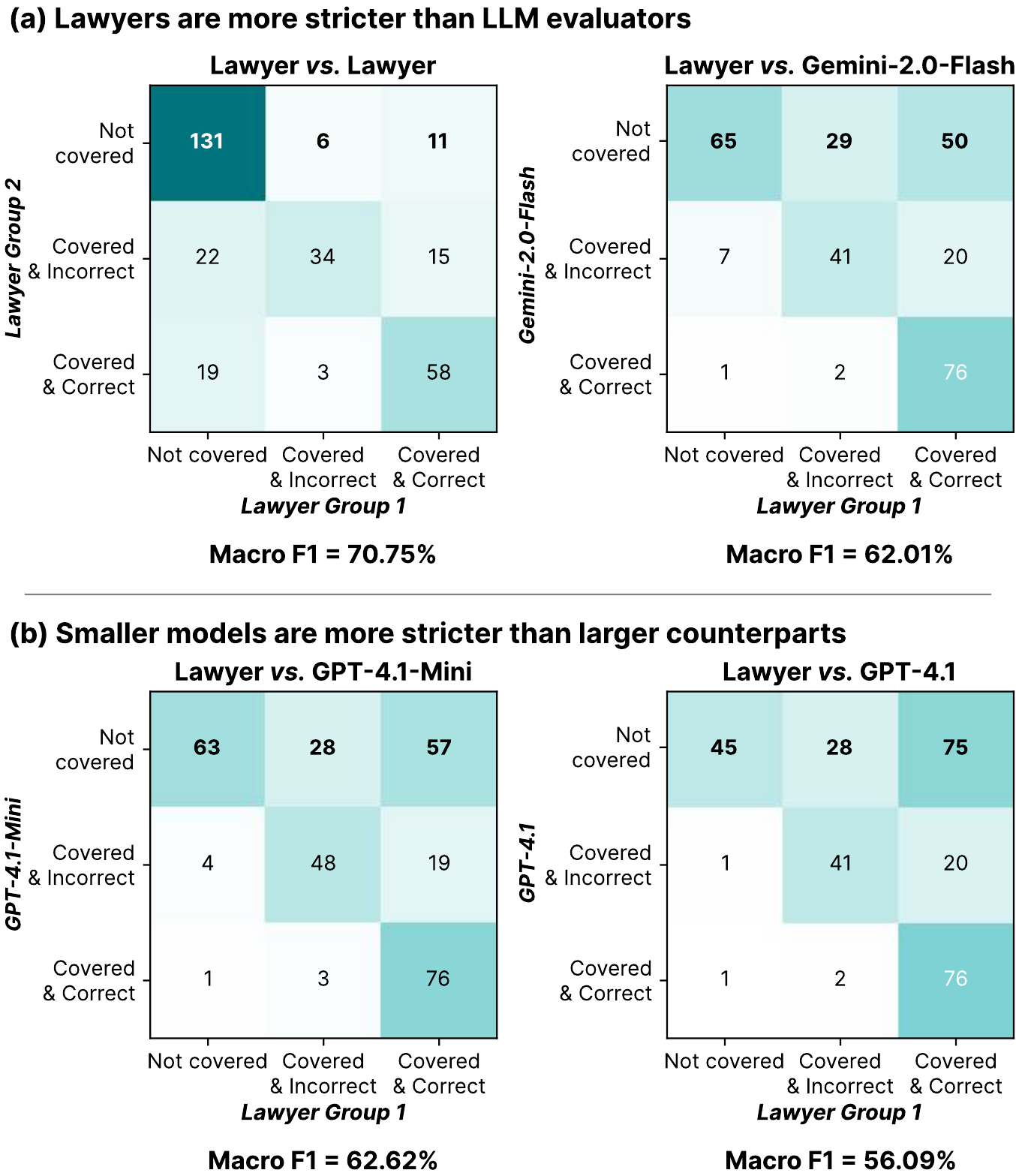}
    \caption{Confusion matrices of individual issue labels between rater groups. (a) Lawyers are the strictest evaluators in judging issue coverage and correctness, and (b) smaller models tend to be stricter than larger counterparts (e.g., GPT-4.1 vs. GPT-4.1-mini, Gemini-2.5-Flash vs. Gemini-2.5-Pro).}
    \label{fig:confusion_matrices}
\end{figure}
\begin{figure*}[t]
    \centering
    \includegraphics[width=\linewidth]{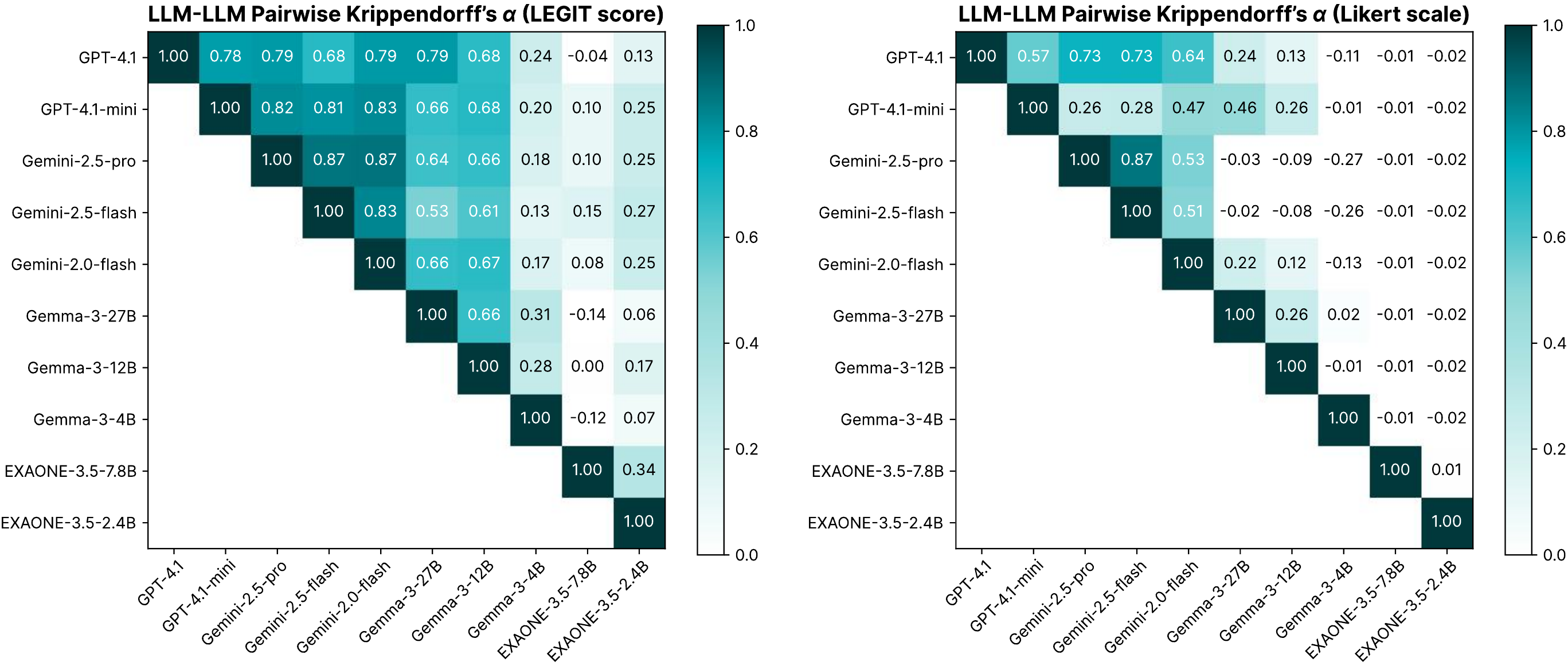}
    \caption{Comparison between LLM-evaluated scores between LEGIT score and Likert scale. Even though the Likert scale prompt includes the ground truth court judgments and rubrics, the coarse granularity limits the inter-rater agreement of LLM-as-a-judge compared to modular LEGIT rubrics.}
    \label{fig:legit_likert_comparison}
\end{figure*}

\paragraph{Error analysis.} We further analyze the discrepancy between lawyers and LLM evaluators. Figure \ref{fig:confusion_matrices} presents two confusion matrices (lawyers \textit{vs.} lawyers, and lawyers \textit{vs.} Gemini-2.0-Flash). Comparison of these matrices shows that the LLM tends to overestimate both issue coverage and correctness, while rarely misclassifying covered or correct issues as uncovered or incorrect. In other words, human evaluators apply stricter standards when assessing the equivalence between legal concepts in the issue description and the reasoning trace, whereas the LLM-as-a-judge adopts looser criteria by mistaking similar but actually different legal concepts. This observation also explains the better lawyer-LLM agreement of smaller models, as they evaluate issues with stricter criteria than their larger counterparts.

\paragraph{} Further details regarding human expert evaluations and LLM prompts are included in Appendix \ref{sec:appendix-lawyerannot} and \ref{sec:appendix-prompts}, respectively.

\subsection{Comparing LEGIT with coarser rubrics}

\subsubsection{Experimental settings}
LLM-as-a-judge is typically applied with the Likert scale \citep{li_generative_2024}, where the evaluator assigns a fixed-range score to the entire response. Multiple studies have proved that these evaluation results are inconsistent across different evaluator models \citep{li_llms-as-judges_2024, han-etal-2025-courtreasoner}. We propose that LEGIT-style issue-based rubrics are more consistent compared to coarse Likert scale rubrics that attempt to evaluate the reasoning trace at once.

As a baseline, we use Likert scale rubrics with an integer score between 0 and 10. To ensure fair comparison with LEGIT, we provide descriptions for intermediate scores (0, 3, 7, 10) regarding issue coverage and correctness, and give the entire judgment text for reference. While it has access to all arguments and the court's conclusions, this rubric is \textit{coarse} in that it evaluates the entire trace within a single inference, and \textit{underspecified} because there is no explicit guidance on what should be included in the trace. We use the same set of generator and evaluator models as in the previous section.

\subsubsection{Results}

\textbf{Modular LEGIT rubrics show higher evaluation consistency than the coarse Likert scale.} Figure \ref{fig:legit_likert_comparison} shows the pairwise LLM-LLM agreement (Krippendorff's $\alpha$) of LEGIT scores and Likert scale evaluations. Even though LLM-as-a-judge for the Likert scale has full information about the case, the modular nature of LEGIT rubrics allows a significant boost in pairwise inter-rater agreement for all LLMs. This shows that LEGIT rubrics are more robust to the selection of the evaluator LLM, likely bringing consistent evaluation results in practice.

\begin{figure}
    \centering
    \includegraphics[width=\linewidth]{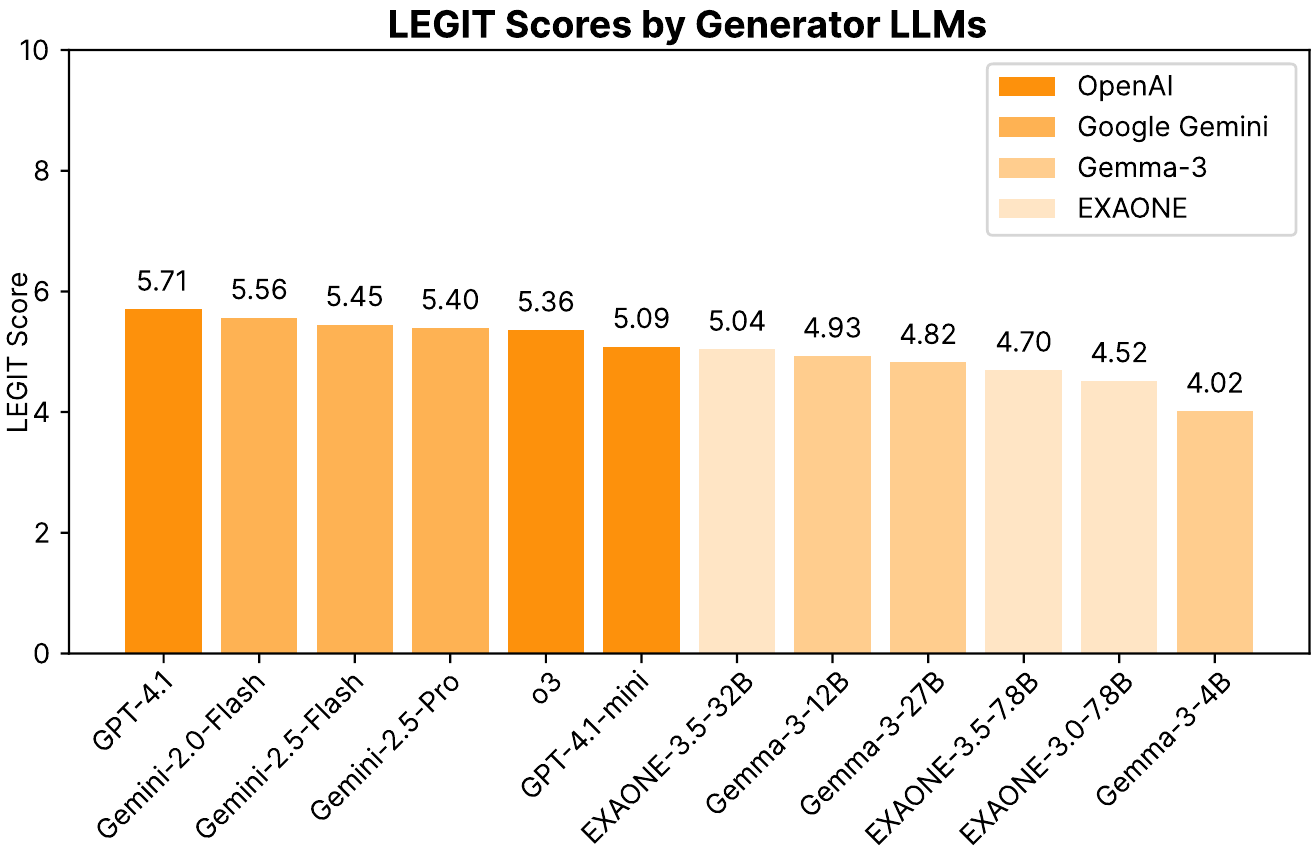}
    \caption{LEGIT score of 12 generator LLMs, evaluated with Gemini-2.0-Flash.}
    \label{fig:legit_score}
\end{figure}

\begin{figure*}
    \centering
    \includegraphics[width=\linewidth]{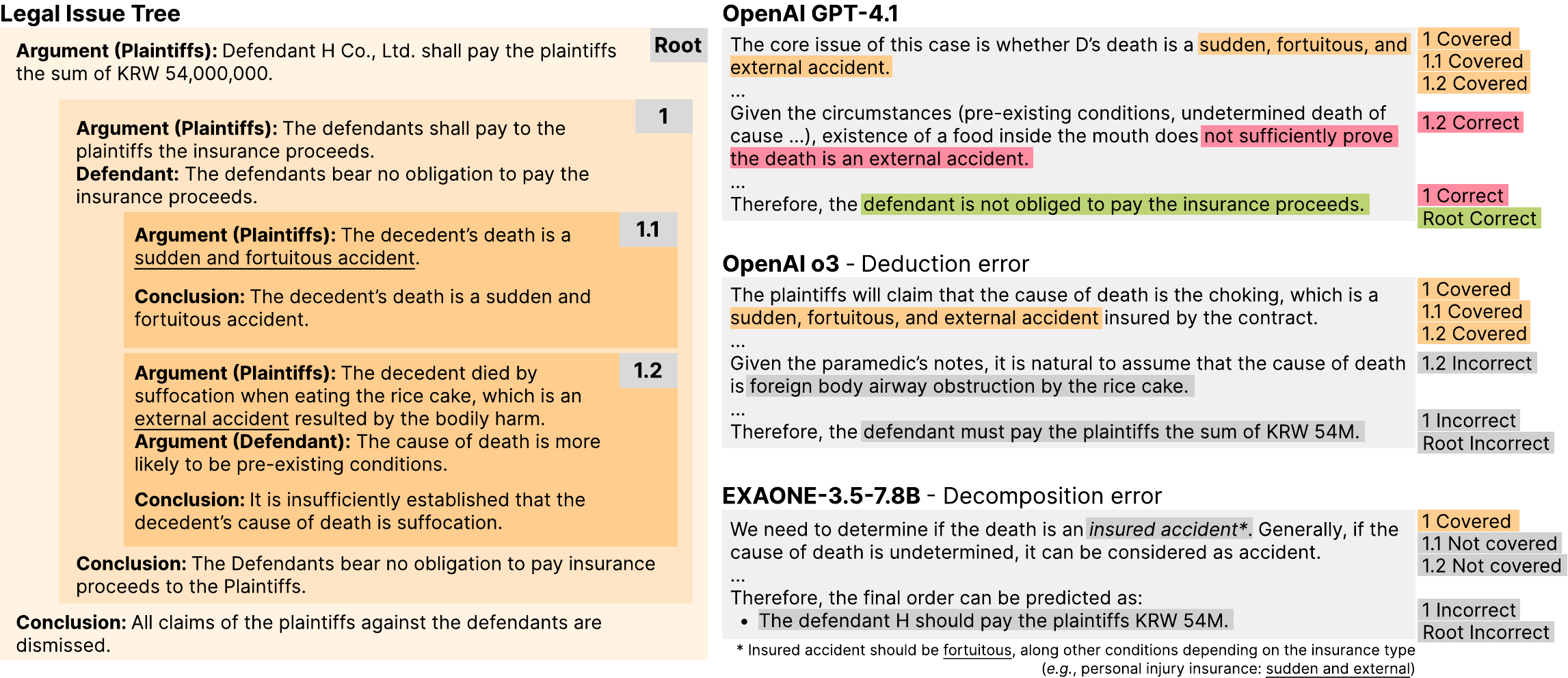}
    \caption{Three LLM responses obtained from the example LEGIT problem in Figure \ref{fig:legit_overview}. o3 fails to reason correctly about \textit{Issue 1.2}, while EXAONE-3.5 does not further decompose \textit{Issue 1} into subissues.}
    \label{fig:qualitative-example}
\end{figure*}

\section{Evaluating LLMs in legal reasoning}
\label{sec:performance}


We now analyze the performance of different LLMs on the LEGIT benchmark. In this section, we use Gemini-2.0-Flash as the evaluator based on its high agreement with human experts (Figure \ref{fig:llm_lawyer_agreement}).

Figure \ref{fig:legit_score} displays the performance of different LLMs. Even the LLMs with strong reasoning capabilities do not saturate the LEGIT task, where the highest LEGIT score is 5.71/10 achieved by GPT-4.1. Since all instances are generated from real court cases and guaranteed to be answerable, this score indicates that even the most powerful LLMs achieve suboptimal performance in complex legal judgment prediction scenarios.

Qualitatively, we identify two types of error, \textbf{deduction error} and \textbf{decomposition error}. During the top-down reasoning process, deduction errors occur when the model fails to reason correctly about complex and conflicting facts, while decomposition errors happen when the LLM fails to identify subissues due to a lack of legal knowledge. Figure \ref{fig:qualitative-example} shows examples of these errors obtained from the example in Figure \ref{fig:legit_overview}. o3 makes a deduction error by incorrectly claiming that the cause of death is choking due to distracting evidence (paramedic report). On the other hand, EXAONE-3.5-7.8B fails to list three conditions of an insured accident defined by law, which led to an irrelevant case analysis and incorrect reasoning.

\begin{figure}[h]
    \centering
    \includegraphics[width=\linewidth]{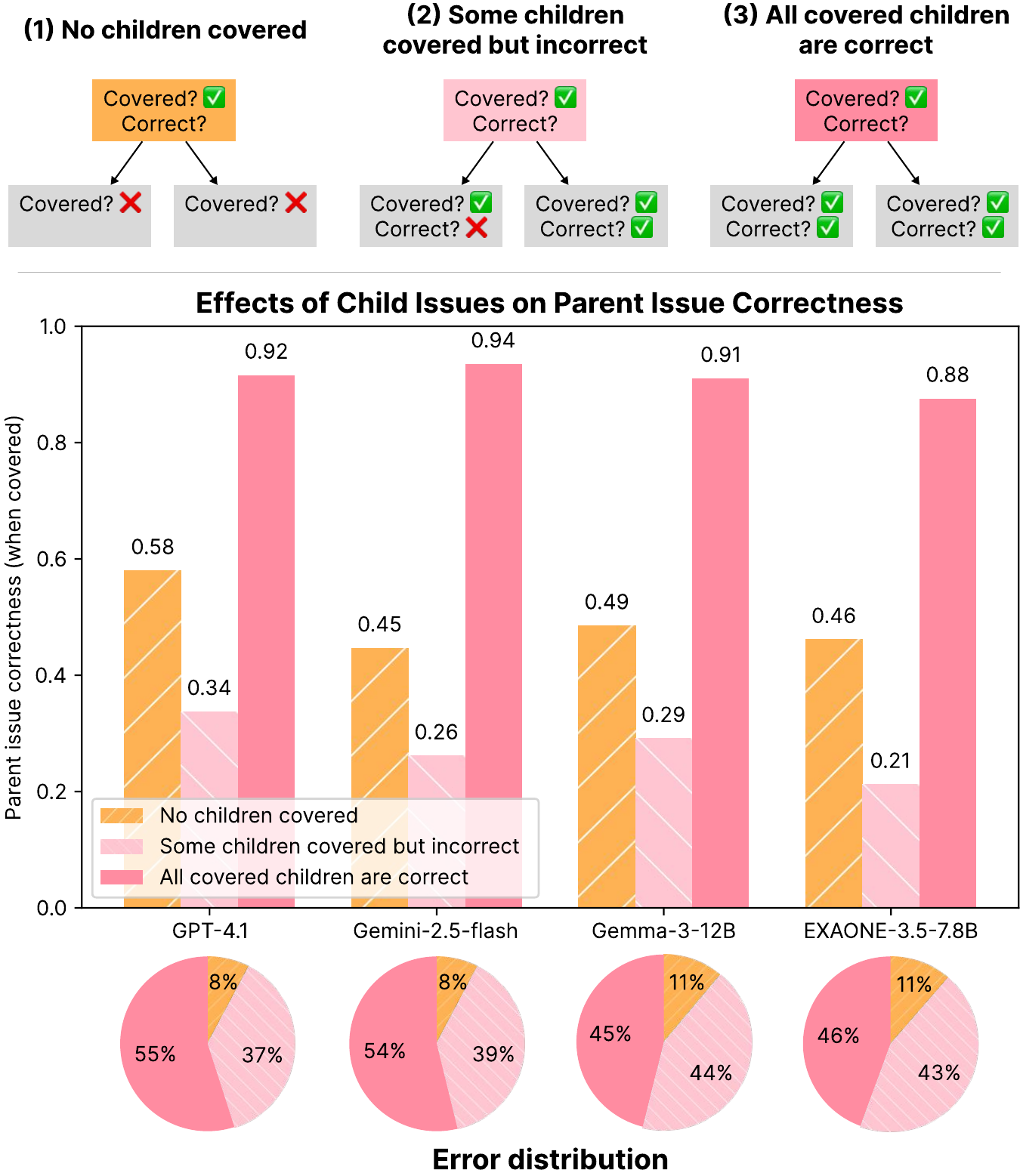}
    \caption{Correctness rate of covered parent issue, depending on child issue results. Failing to identify child issues or to reason about them correctly seriously degrades the correctness of parent issues.}
    \label{fig:effects_of_children_node_results}
\end{figure}

We perform a quantitative analysis of how deduction and decomposition errors affect the reasoning performance. Specifically, for each depth-1 subtree of LEGIT, we measure the parent issue's correctness when (1) no child issues are covered (decomposition error), (2) at least one child issue is covered but incorrect (deduction error), and (3) all covered child issues are correct. Compared to the ideal case (3), both errors cause serious degradation in the parent issue's correctness (Figure \ref{fig:effects_of_children_node_results}). In other words, when LLMs fail to identify the subissues or their reasoning on the child issues is incorrect, these errors are likely to propagate to the higher issues, leading to incorrect reasoning traces and final order prediction. Hence, it is crucial to reduce deduction and decomposition errors by improving issue coverage and correctness in legal reasoning.

Appendix \ref{sec:appendix-quantiative} includes further quantitative results about how problem difficulty, reasoning trace quality, and final order correctness relate.



\section{Improving legal reasoning capabilities}
\label{sec:improving}


As shown in the preceding analysis, low issue coverage and correctness can negatively affect the overall reasoning performance. In this section, we explore two of the most commonly used approaches for augmenting the legal reasoning capability of LLMs, retrieval-augmented generation (RAG) and reinforcement learning (RL). We test a minimal variant of each method to isolate and understand its effects on the LEGIT score. 

\subsection{Experimental settings}

\paragraph{Retrieval-augmented generation (RAG)} RAG can increase the factuality of LLMs by retrieving and prepending relevant documents to the input \citep{lewis_retrieval-augmented_2020, gao_retrieval-augmented_2024}. In the context of legal RAG, we explore \textit{citation retrieval} \citep{zhang_citalaw_2025}, a challenging form of \textit{retrieval for reasoning} \citep{su_bright_2025, shao_reasonir_2025} where the retriever should search for relevant statutes and leading cases to support the reasoning process. We extract all citations from the LEGIT dataset (both train and test splits) as the retrieval base, and use the fact and purpose of claim as the query.

We employ lexical matching-based BM25 \citep{robertson_simple_1994} and dense vector-based mContriever \citep{izacard_unsupervised_2022}, two commonly used retrievers for legal RAG \citep{rosa_yes_2021, kim_legalsearchlm_2025}. We also test RAG with ground truth (GT) citations as an ideal case. Ten retrieved citations obtained from each retriever (all citations for GT) are prepended to the original LEGIT problem and passed to Gemma-3-4B. We again use the same evaluator model (Gemini-2.0-Flash) for evaluation.

\paragraph{Reinforcement learning (RL) with LEGIT rubrics} RL provides a way to directly optimize LLMs for any specified metric  \citep{shen_minimum_2016, ouyang_training_2022}, including rubric-based scores \citep{viswanathan_checklists_2025, gunjal_rubrics_2025}. To improve its legal reasoning ability, we train a Gemma-3-4B checkpoint using the GRPO objective \citep{shao_deepseekmath_2024} using LEGIT scores as rewards. We use Gemma-3-27B as the evaluator during the training phase, and Gemini-2.0-flash for evaluation at test time. Using separate LLM-as-a-judge models for training and testing improves evaluation robustness by preventing the policy from overfitting to the training-time evaluator.




\paragraph{} Refer to Appendix \ref{sec:appendix-rag}-\ref{sec:appendix-rl} for preprocessing, hyperparameters, computation budgets, and additional experiments (fine-tuned Contriever and more generators for RAG, final answer-only rewards for RL).

\subsection{Results}

\begin{figure}
    \centering
    \includegraphics[width=\linewidth]{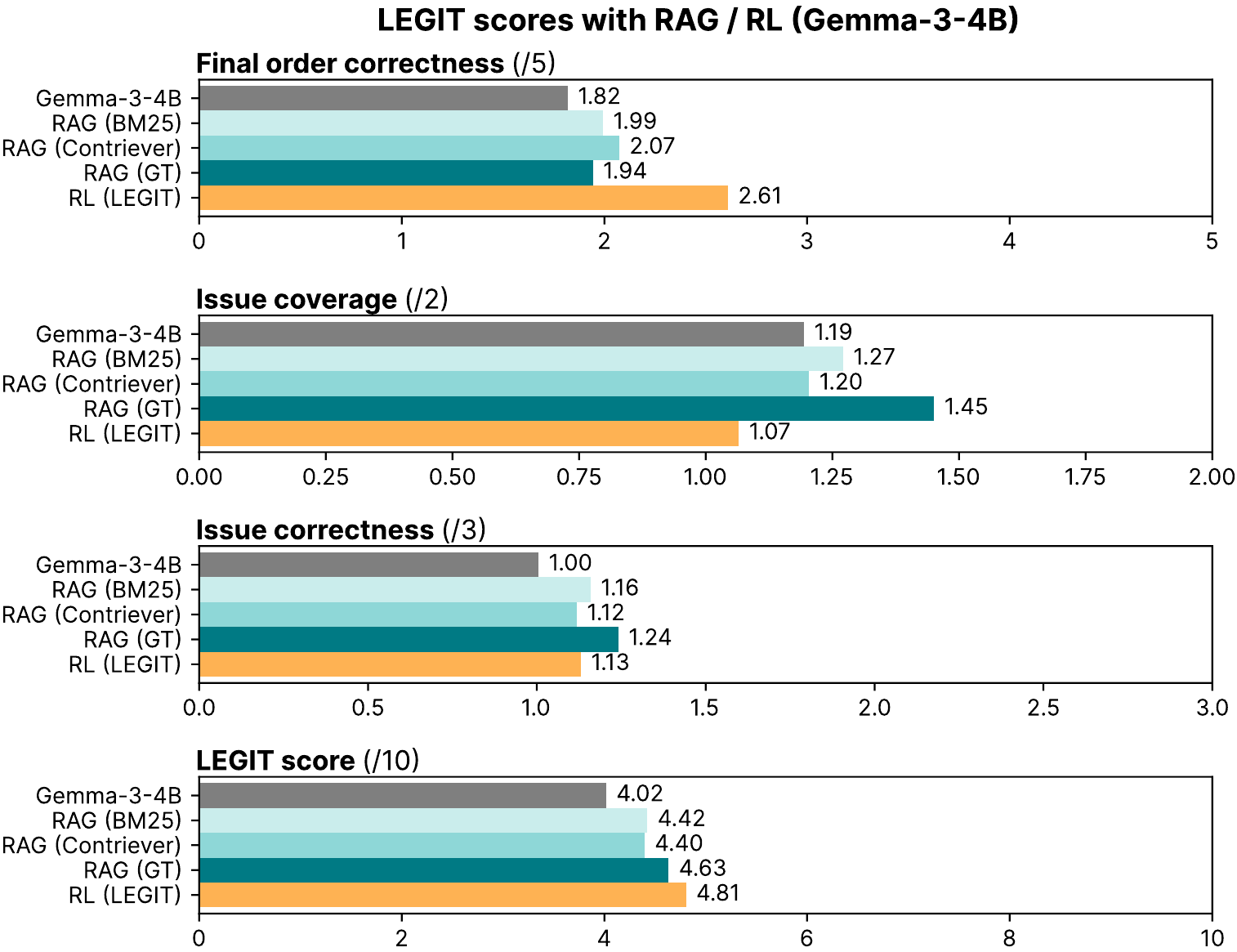}
    \caption{Comparison of LEGIT scores of Gemma-3-4B with RAG and RL. While RAG improves all components of LEGIT scores, RL significantly improves the final order/issue correctness while \textit{reducing} issue coverage.}
    \label{fig:legit_score_rag_rl}
\end{figure}

We analyze the effects of RAG and RL by comparing their impact on the three components of the LEGIT score (Figure \ref{fig:legit_score_rag_rl}).

\paragraph{RAG improves all reasoning abilities.} Prepending citations retrieved by BM25 or Contriever improves LEGIT scores by around +0.4 points. As shown in Figure \ref{fig:legit_score_rag_rl}, these improvements are distributed across all three components, as retrieved laws directly indicate possible arguments and serve as an inference rule during reasoning. Closed-sourced LLMs show consistent results; see Appendix \ref{sec:appendix-rag}.


\paragraph{RL prioritizes correctness at the cost of coverage.} In contrast to RAG, reinforcement learning with the LEGIT reward significantly increases final order and issue correctness, but \textit{reduces} issue coverage. This trade-off is consistent with the analysis in Figure \ref{fig:effects_of_children_node_results}, which shows that the penalty for incorrect reasoning is more severe than omitting the issue altogether. As a result, the policy trained with RL will likely favor covering only those issues that are straightforward, while avoiding fuzzier or subtler issues that might negatively affect the parent nodes.

\paragraph{}Overall, these results highlight the complementary effects of RAG and RL. RAG allows broader exploration and accurate reasoning by providing relevant law, while RL sharpens the model’s reasoning correctness by pruning uncertain issues. Therefore, combining the two approaches holds promise for improving the legal reasoning ability of LLMs.
\section{Conclusion}

This work proposes LEGIT, a high-quality dataset of LJP problems and legal issue tree-based rubrics for reasoning traces. LEGIT's rubrics allow reliable and consistent LLM-as-a-judge evaluation, achieving strong agreement with human experts. Furthermore, we show that LLMs often fail to identify relevant issues or reason about them correctly, which affects the correctness of higher-level issues and harms the response quality. Finally, we show that {RAG} and {RL with LEGIT rubrics} have complementary benefits, where RAG benefits general reasoning ability while LEGIT reward improves correctness by reducing issue coverage.

Evaluating reasoning traces is crucial for developing AI for high-stakes domains like law. Structured rubrics extracted from court judgments allow reliable evaluation of issue correctness and coverage, provide insights about how different failures propagate through the issue hierarchy, and can directly improve LLMs via RL. We believe our work represents an important step toward developing expert-level reasoning LLMs.

\section{Limitations}

First, the proposed LEGIT dataset only addresses the Korean legal system and is limited to the Korean language. However, we believe that this work has the potential to generalize beyond a single legal system and language. A proposal for evaluating and improving issue coverage of LLM legal reasoning can be found in multiple previous works across different jurisdictions \citep{izzidien_llm_2024, yu_structured_2025}. Furthermore, we believe that our findings on LLMs' performance in legal reasoning (Section \ref{sec:performance}) and how RAG and RL improve it (Section \ref{sec:improving}) can be generalized beyond the data scope of this work. We believe extending LEGIT to diverse jurisdictions and languages is a promising direction, and we leave it as future work.

Furthermore, LEGIT's rubric-based evaluation requires more compute than other LLM-as-a-judge methods, \textit{i.e.}, Likert scale or evaluating only final order accuracy. We view this as a tradeoff between computation and evaluation reliability, echoing works from other fields that show scaling evaluation compute leads to better evaluation performance \citep{hashemi_llm-rubric_2024, lee_checkeval_2025, kim_scaling_2025}.

Finally, LEGIT does not directly address \textit{citation accuracy}, evaluating whether the LLM cites the correct source document for the quoted legal knowledge. Such a decision was intentional, as we frequently observe false negatives when evaluating citation accuracy. For instance, we identify at least 25 different Supreme Court case identifiers that are associated with the same case law (about claims that can be protected by the rights to revoke fraudulent conveyances) in the LEGIT dataset, while each case judgments cite only one or two of them. To properly handle such cases, we need to verify whether the cited document actually exists and is relevant given the context \citep{byun_this_2024, wu_automated_2025}. However, as Korean court judgments are not freely disclosed to the public, we find this approach infeasible at the time of writing.

\newpage

\bibliography{references}

\newpage

\appendix

\section{Additional LEGIT example}
\label{sec:appendix-example}

\begin{figure*}
    \centering
    \includegraphics[width=\linewidth]{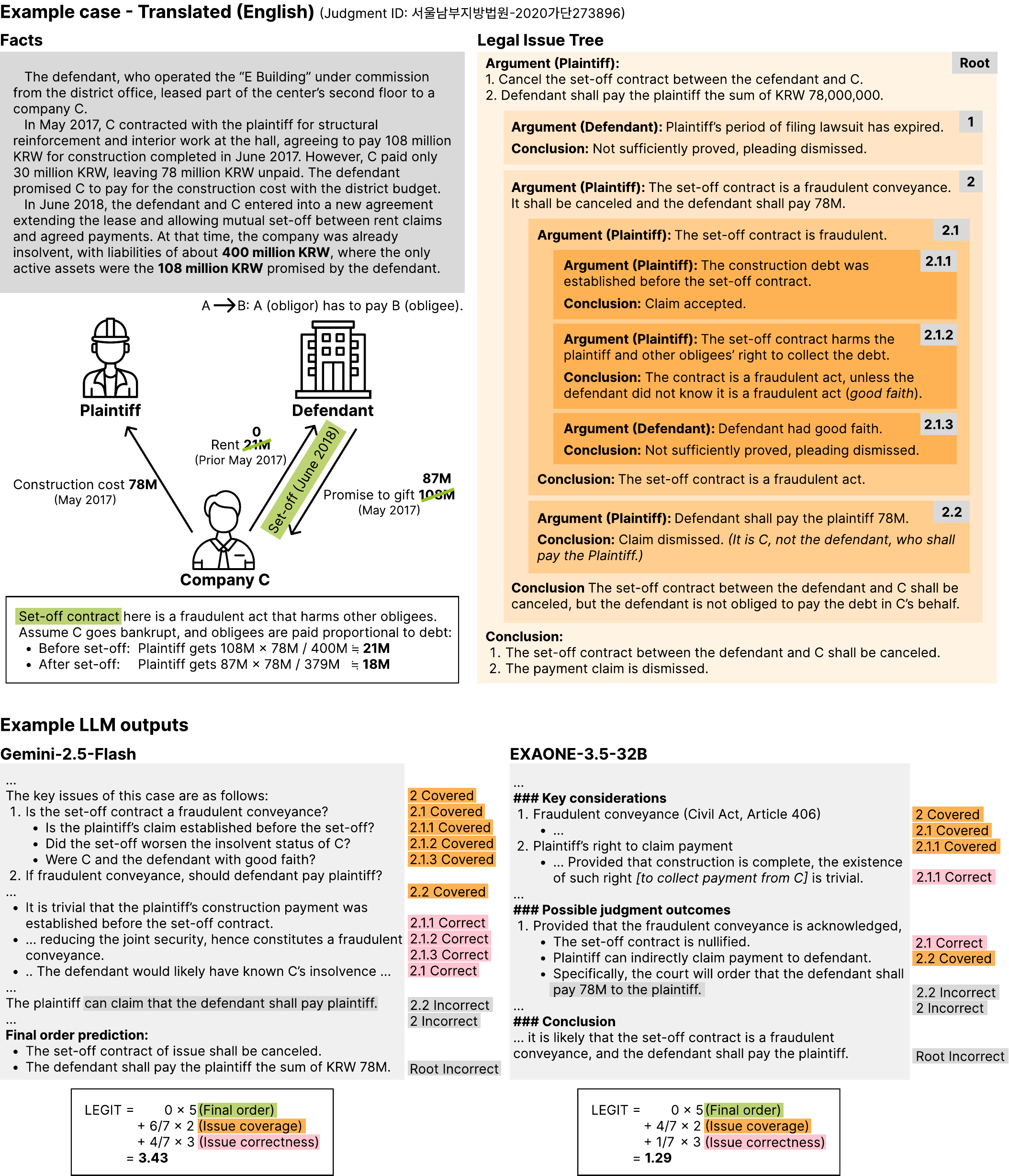}
    \caption{A detailed example of a LEGIT case (fraudulent conveyance, top), including facts and the legal issue tree, as well as two LLM outputs and their LEGIT scores (bottom),  translated into English. Refer to Figure \ref{fig:detailed_example_korean} for the original version of the data and LLM responses.}
    \label{fig:detailed_example}
\end{figure*}

\begin{figure*}
    \centering
    \includegraphics[width=\linewidth]{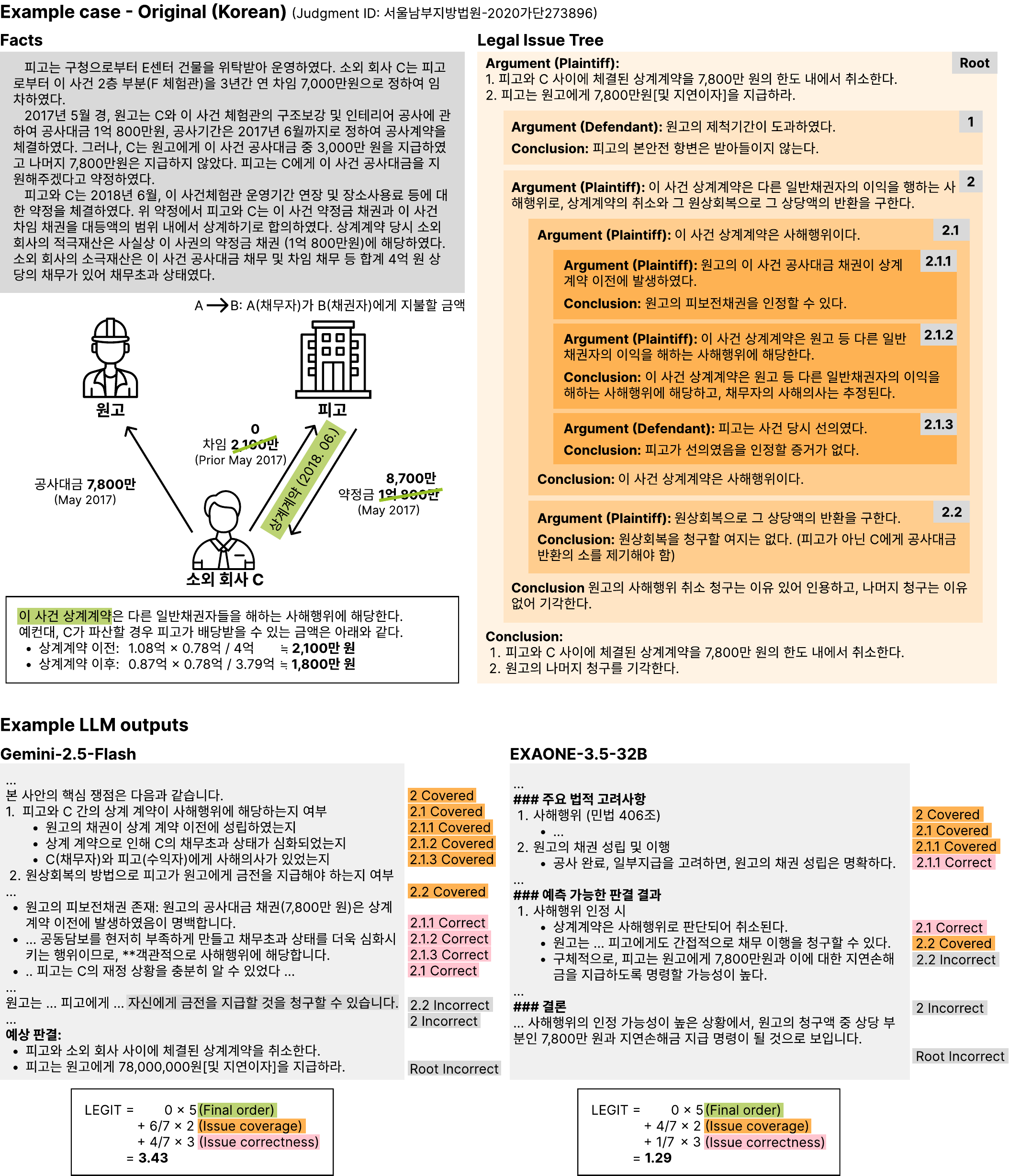}
    \caption{A detailed example of a LEGIT case (fraudulent conveyance, top), including facts and the legal issue tree, as well as two LLM outputs and their LEGIT scores (bottom), in Korean. Refer to Figure \ref{fig:detailed_example} for the English-translated version of the data and LLM responses.}
    \label{fig:detailed_example_korean}
\end{figure*}

As another representative (and more complex) example of LEGIT, we show a \textit{fraudulent conveyance} case and responses obtained from two LLMs, Gemini-2.5-Flash and EXAONE-3.5-32B (see Figure~\ref{fig:detailed_example} for the English version, and Figure~\ref{fig:detailed_example_korean} for Korean).

In the Korean Civil Act, a fraudulent conveyance is when the obligor (someone in debt) makes a contract with a third party (\textit{e.g.}, selling/giving their property, setting off debt, etc.), knowing that it will harm the obligee's rights to collect the debt back. Within a year of when the obligee acknowledged that a fraudulent conveyance had happened, the obligee can sue the third party to cancel the fraudulent contract. In this particular scenario, the obligor (company C) pays off the debt of only one obligee (defendant), which reduces the joint security for other obligees and harms their right to reclaim debt.

\begin{quote}
   \textbf{Article 406 (Obligee's Right of Revocation)} \\
   (1) If the obligor has performed any juristic act which has a property right for its subject, with the knowledge that it would prejudice the obligee, the obligee may apply to the court for its revocation and restitution of its original status: Provided, That this shall not apply where a person who has derived a benefit from such act or a subsequent purchaser was, at the time of the act or of the purchase, unaware of the fact that it would prejudice the obligee.
\end{quote}

\begin{figure*}[t]
    \centering
    \includegraphics[width=0.8\linewidth]{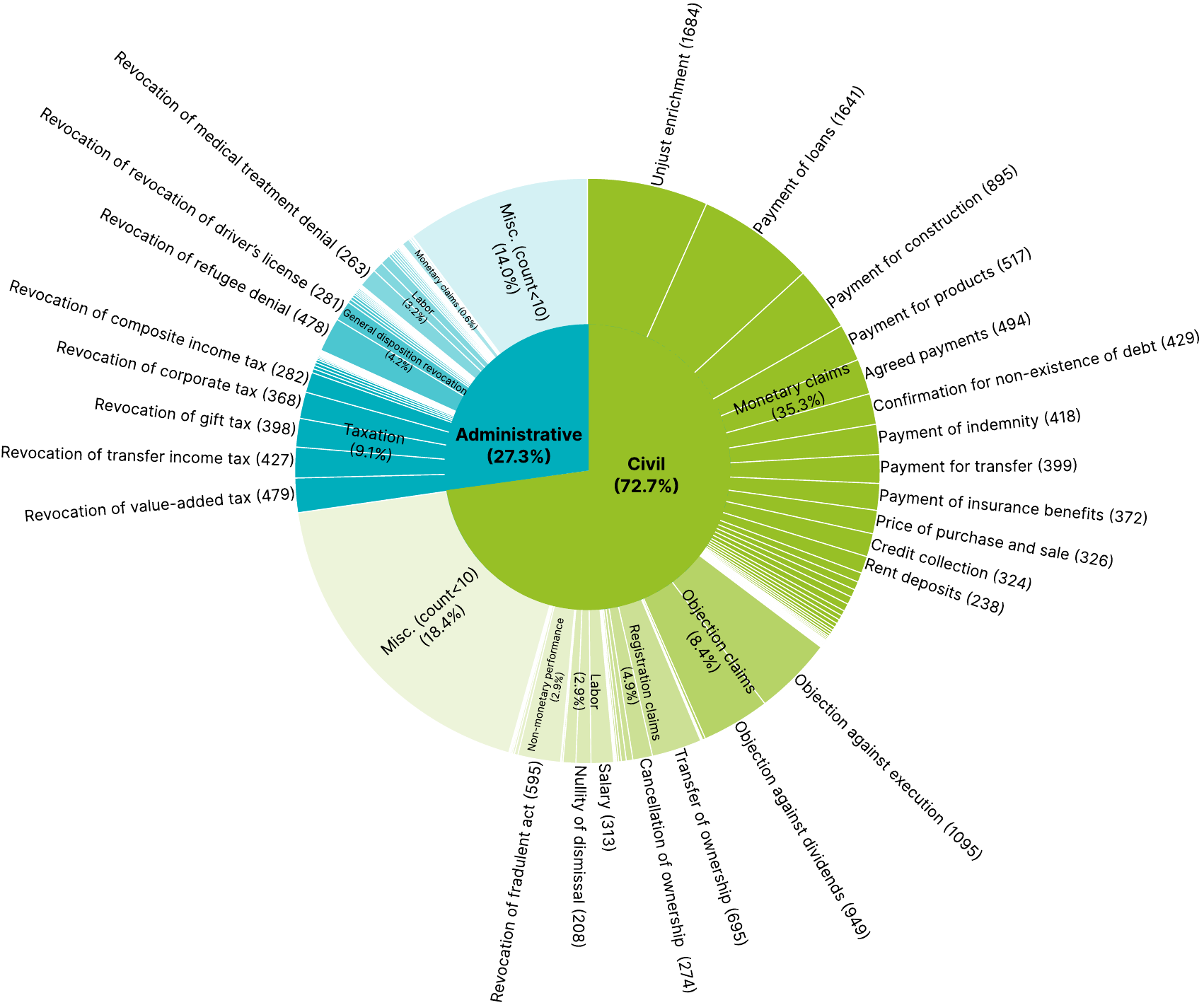}
    \caption{Distribution of case types in LEGIT. Case types that have more than 200 instances are shown with their instance count. \textit{Misc.} subcategory includes all case types with under 10 instances. Compared to other LJP benchmarks, LEGIT includes an unprecedented variety of civil and administrative cases.}
    \label{fig:legit_casetype_distribution}
\end{figure*}

\begin{figure}[t]
    \centering
    \includegraphics[width=\linewidth]{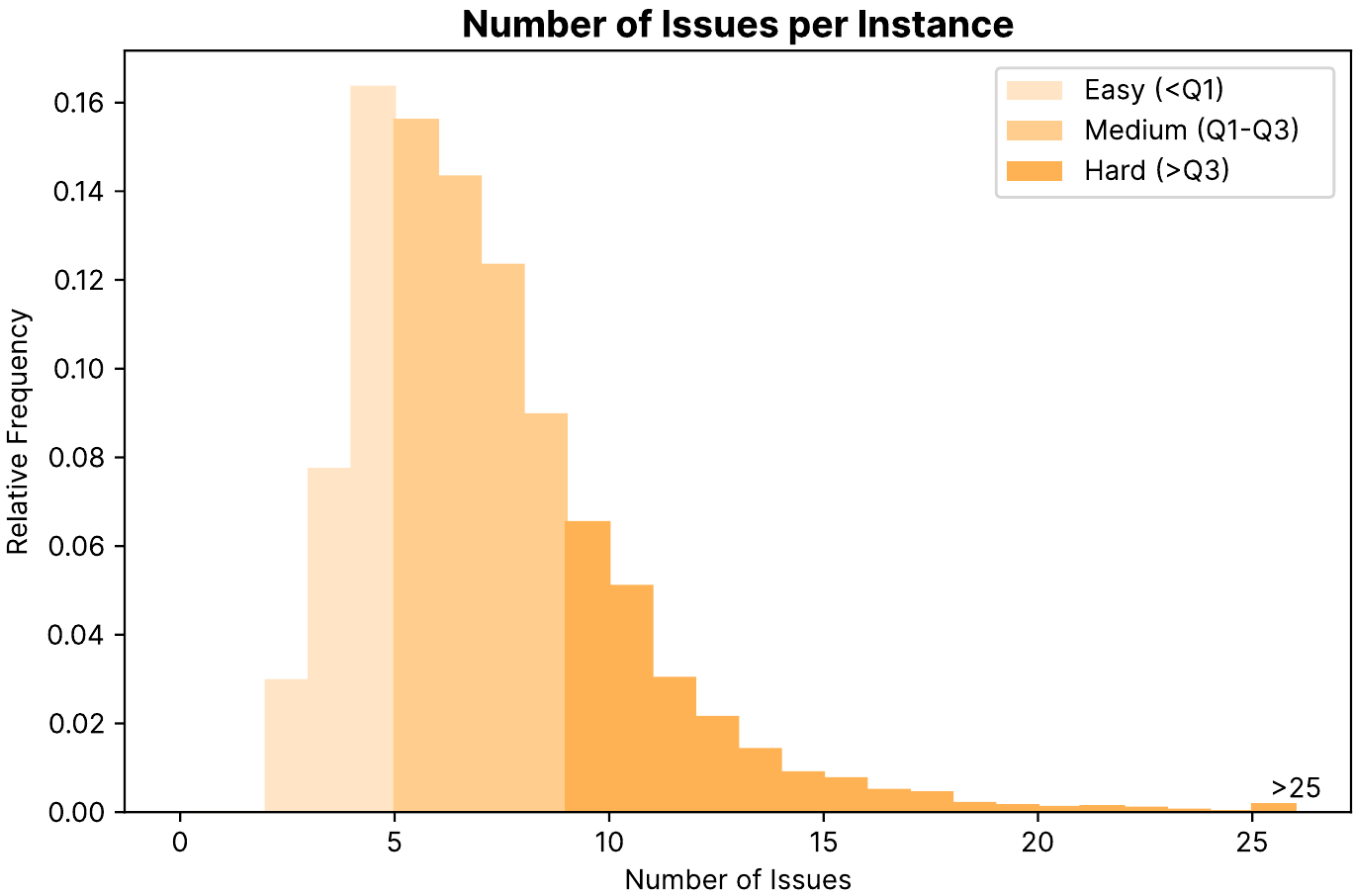}
    \caption{Histogram showing the number of issues for each LEGIT instance. The dataset is divided into easy/medium/hard difficulty subsets based on the number of issues.}
    \label{fig:legit_issue_count}
\end{figure}

\begin{table*}[tb]
    \centering
    \footnotesize
    \begin{tabular}{p{3.7cm}p{9cm}p{1.5cm}}
        \toprule
        \textbf{Error types} & \textbf{Description} & \textbf{Occurence} \\
        \midrule
        \multicolumn{3}{l}{\textbf{Fact extraction} (/50 facts)} \\
        Missing antecedents & The antecedents of some pronouns and redacted symbols are not directly mentioned (can be inferred by context). & 14.0\% \\
        Overspecification & Fact paragraph includes information that is not a neutral statement. & 10.0\% \\
        Missing numbers$^\dagger$ & Important numbers (\textit{e.g.}, money) are redacted in the raw data, affecting multiple issues. & 6.0\% \\
        Facts in attachments* & A major portion of the facts are included in the attachments, not in the original judgment. & 2.0\% \\
        \midrule
        \multicolumn{3}{l}{\textbf{Issue structure extraction} (/306 issues)} \\
        Ungrounded claims & Parties' claims are not grounded in the facts, thus rejected by judges due to a complete lack of evidence. & 5.2\% \\
        Vacuous issues & No meaningful claim and conclusion are presented. & 0.7\% \\
        Duplicate issues & There are two issues with identical claim and conclusion. & 0.3\% \\
        \midrule
        \multicolumn{3}{l}{\textbf{Overall answerability} (/50)} \\
        Answerable & & 92\% \\
        Partially answerable($^\dagger$) & & 6\% \\
        Unanswerable($^*$) & & 2\% \\
        \bottomrule
    \end{tabular}
    \caption{Dataset error types and rates were analyzed from a randomly sampled subset of LEGIT’s training split (50 examples, 306 issues). Dagger ($^\dagger$) represents the case where it is impossible to deduce a significant amount of the issues (>25\%). Asterisks ($^*$) denote critical errors that render a question unanswerable. All such cases stem from defects in the raw data. Other errors are minor and do not prevent experts from correctly predicting the final order or identifying the relevant issues.}
    \label{tab:legit-dataset-errors}
\end{table*}

Although both LLMs make an incorrect final order prediction, the LEGIT rubrics clearly distinguish between the two models, since the LEGIT issue coverage and correctness scores provide richer signals than the final order-only evaluation of traditional LJP benchmarks. First, Gemini is fully aware of the three conditions to declare fraudulent conveyance (identifying issues 2.1.1-2.1.3), correctly reasoning that the set-off contract of this case is a fraudulent conveyance. EXAONE also reaches the conclusion that the set-off is fraudulent, but only points out one of the three conditions (issue 2.2.1). Here, the LEGIT rubrics clearly indicate that Gemini's response is more accurate and informative. Second, both models incorrectly conclude that the defendant shall directly pay the plaintiff (issue 2.2), which pinpoints the deduction error that led to the wrong final order prediction. 

\section{LEGIT dataset details}
\label{sec:appendix-dataset}

\subsection{Filtering out non-deterministic cases}
\label{sec:appendix-dataset-preproc}
We use rules to filter out non-deterministic final orders. First, we only maintain civil and administrative cases, which can be identified by the document ID. Then, we filter out any judgments that include the following keywords curated by a lawyer: "compensation for damages (손해배상)", "negligence rate (과실비율)", "liability rate (책임비율)", and "compensation for pain and suffering (위자료)". Cases that include these terminologies are likely to involve the judge's discretion, which we do not allow in the LEGIT dataset.

\subsection{Dataset statistics}
\label{sec:appendix-dataset-stats}

\paragraph{Issue count} The number of issues indicates the logical complexity of a case. The issue count distribution of the entire LEGIT dataset is shown in Figure \ref{fig:legit_issue_count}. The median issue count is 7, indicating that most cases in the LEGIT dataset carry a complex, nontrivial set of legal issues.

\paragraph{Case types} In Korea, plaintiffs can assign \textit{case types} when filing the lawsuit to indicate the nature of the case, \textit{e.g.}, claim, payment of loans, revocation of corporate tax, etc. While there are no predefined lists of case type identifiers, there are many common case types that are shared by the court and legal practitioners. Note that one judgment can have multiple case types, which occurs when the defendant files a counterclaim or two cases are merged during the trial.
Figure \ref{fig:legit_casetype_distribution} shows the distribution of case types in LEGIT, after applying string regularization. Among 3,697 distinct case types in LEGIT, there are 27 types with more than 200 instances, and 111 types with more than 10 instances. This diversity in civil/administrative cases is unrivaled by existing legal judgment prediction datasets, which often focus on either criminal cases or very narrow subsets of civil cases.

\subsection{Training set quality inspection}

The analysis in Section \ref{sec:reliability} shows that lawyers achieve significant agreement with the extracted rubrics, proving the quality of the test split of the LEGIT dataset. However, unlike the test split, where the authors manually corrected errors in the inputs and rubrics, the training set remains unmodified from the automatically labeled version. Here, we report manual inspection results on the training set.

To assess the quality of extracted facts and legal issue trees, we manually inspect a small subset of the training split. The error types and respective error rates observed from 50 randomly sampled training problems are shown in Table \ref{tab:legit-dataset-errors}. Overall, only one problem was not answerable due to missing information in the raw data, and all errors made in the LLM-based annotation stage were minor since the experts could still deduce the correct final order and identify relevant legal issues. This shows that the LEGIT dataset's automatic annotation is of reasonable quality, which justifies the rubric-based RL.

\section{Expert annotation details}
\label{sec:appendix-lawyerannot}

\begin{figure*}[t]
    \centering
    \includegraphics[width=\linewidth]{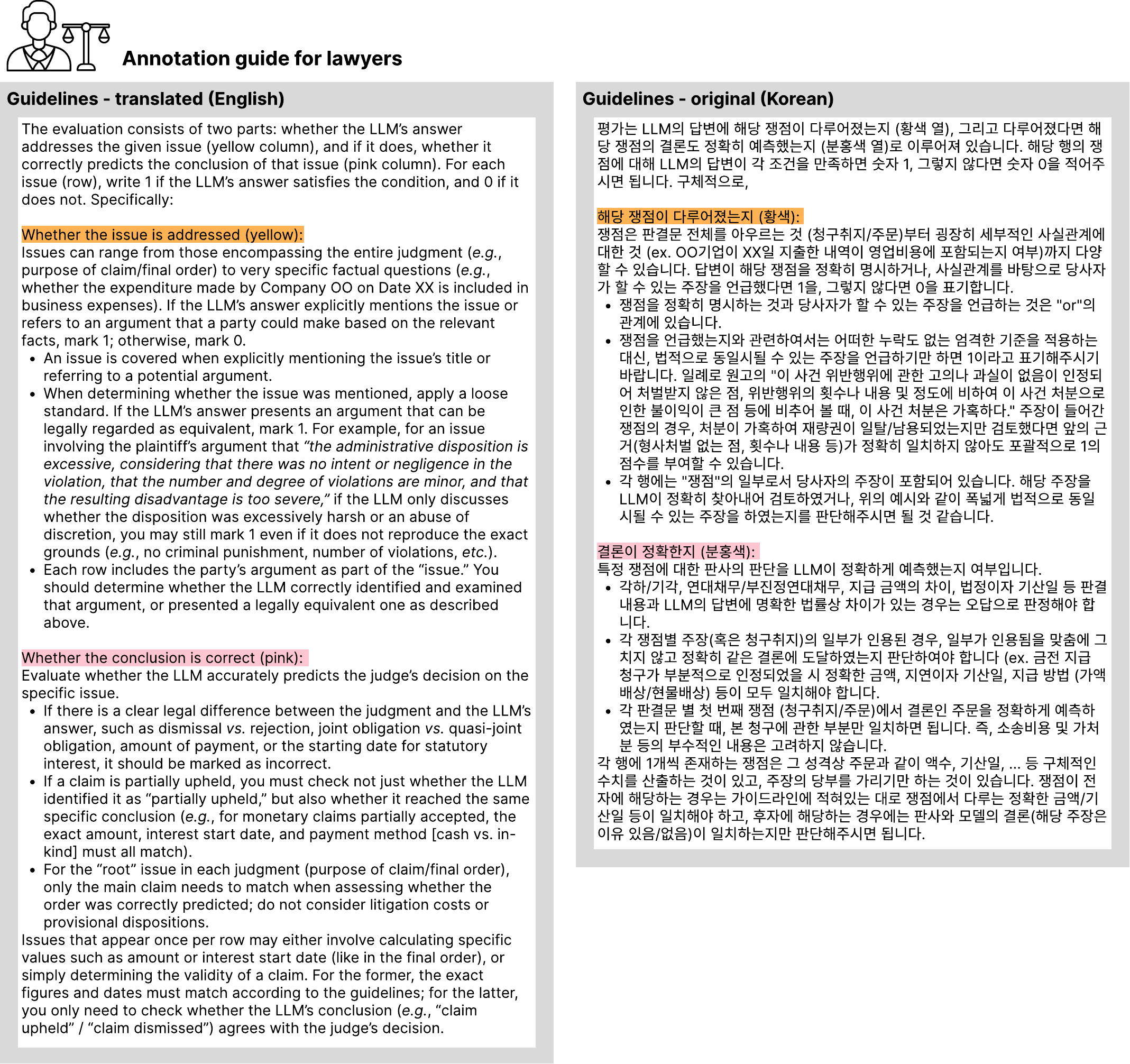}
    \caption{Annotation guide presented to the lawyers during expert annotation process in Section \ref{sec:reliability}.}
    \label{fig:lawyer_anntoation_guide}
\end{figure*}

In Section \ref{sec:reliability}, we collect human expert annotations to assess the quality of LEGIT rubrics and the reliability of LLM-as-a-judge methods. For this process, we hire 7 licensed Korean lawyers with sufficient knowledge in civil and administrative law. To measure inter-human agreement, we divide the lawyers into two groups and instruct them to annotate the same samples, while disallowing communication between groups during the annotation process.

The lawyers are instructed to evaluate LLM-generated reasoning traces obtained from 44 different problems using LEGIT rubrics. The sample set contains 15 easy/15 medium/14 hard problems, so that the total number of issues sums up to 300. For each response, annotators are provided (problem, LLM response, issue) tuples as input, and were asked if the LLM response covered the issue and predicted the conclusion correctly. Note that the instruction did not ask to evaluate the overall quality of the reasoning trace, nor were the lawyers knowledgeable about how the final LEGIT score was computed from their annotation. The full prompt can be seen in Figure \ref{fig:lawyer_anntoation_guide}.

The hourly compensation was set to KRW 264,000 (approx. USD 187), and the total work time of seven lawyers is 12.78 hours.

\newpage

\section{Additional quantitative analyses}
\label{sec:appendix-quantiative}

\begin{figure}[t]
    \centering
    \includegraphics[width=\linewidth]{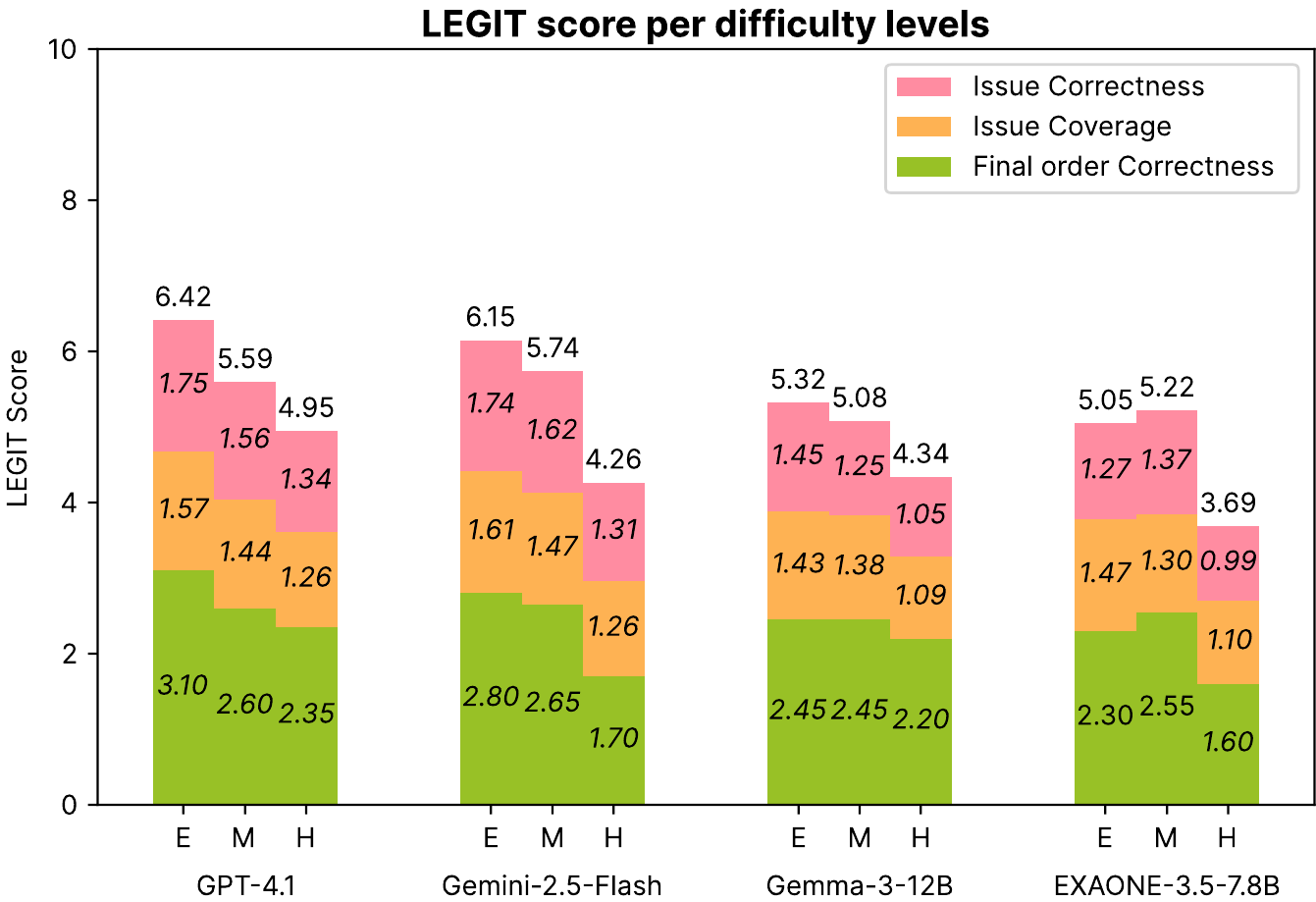}
    \caption{Component-wise LEGIT score of four LLMs, divided by difficulty subsets (E: Easy, M: Medium, H: Hard). Individual score components (final order accuracy, issue coverage, issue correctness) generally drop as the case becomes more complex.}
    \label{fig:difficulty}
\end{figure}

\begin{figure}[t]
    \centering
    \includegraphics[width=\linewidth]{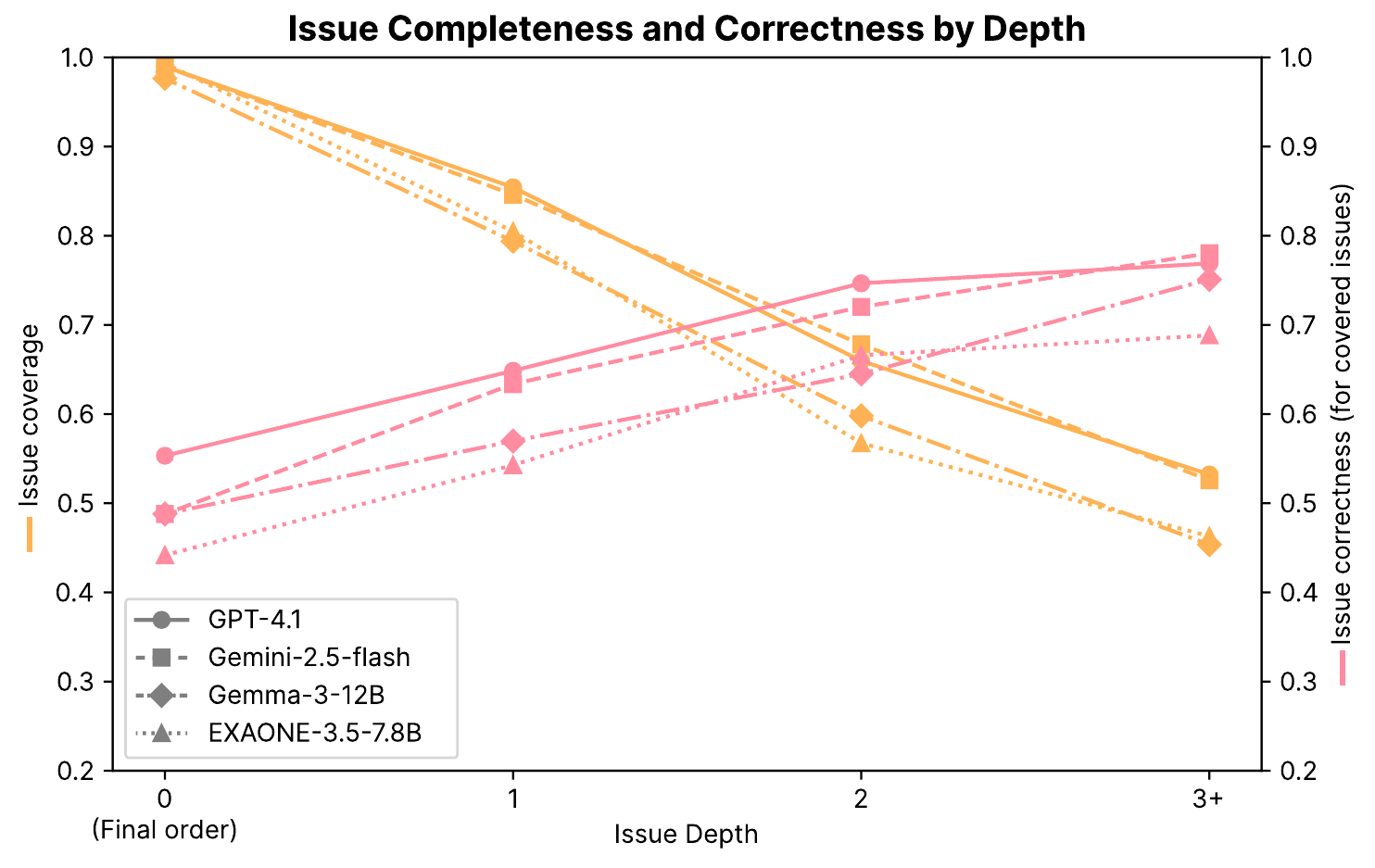}
    \caption{(1) Issue coverage and (2) issue correctness of covered issues ($\frac{\mathrm{correct}}{\mathrm{covered}}$) per issue depth.}
    \label{fig:accuracy_per_depth}
\end{figure}

\begin{figure}[t]
    \centering
    \includegraphics[width=\linewidth]{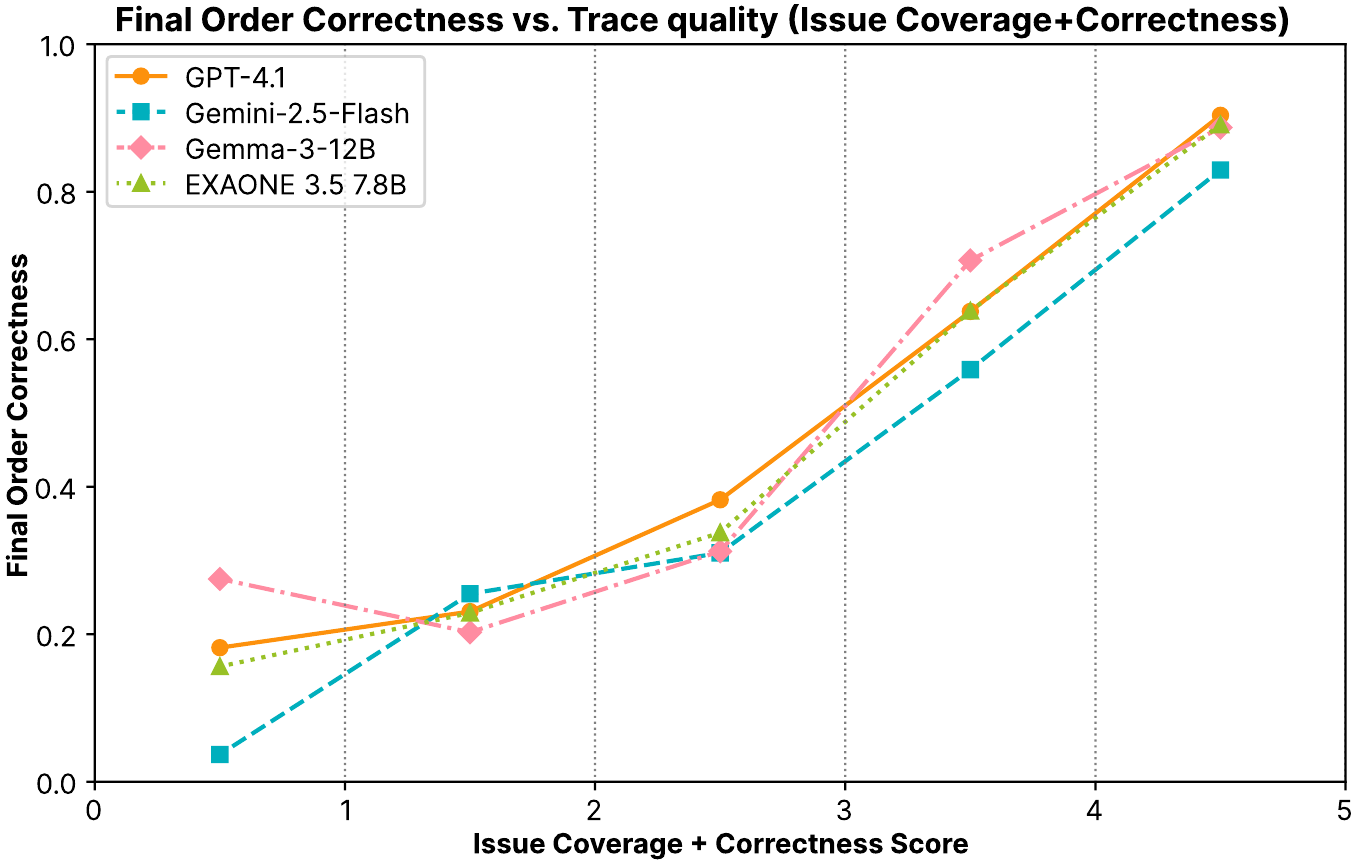}
    \caption{Issue coverage + correctness scores strongly correlate with final order correctness, suggesting that reasoning trace quality has a \textit{causal} effect on the final answer accuracy.}
    \label{fig:trace_quality_and_final_order_corr}
\end{figure}

\begin{figure*}[t]
    \centering
    \includegraphics[width=0.8\linewidth]{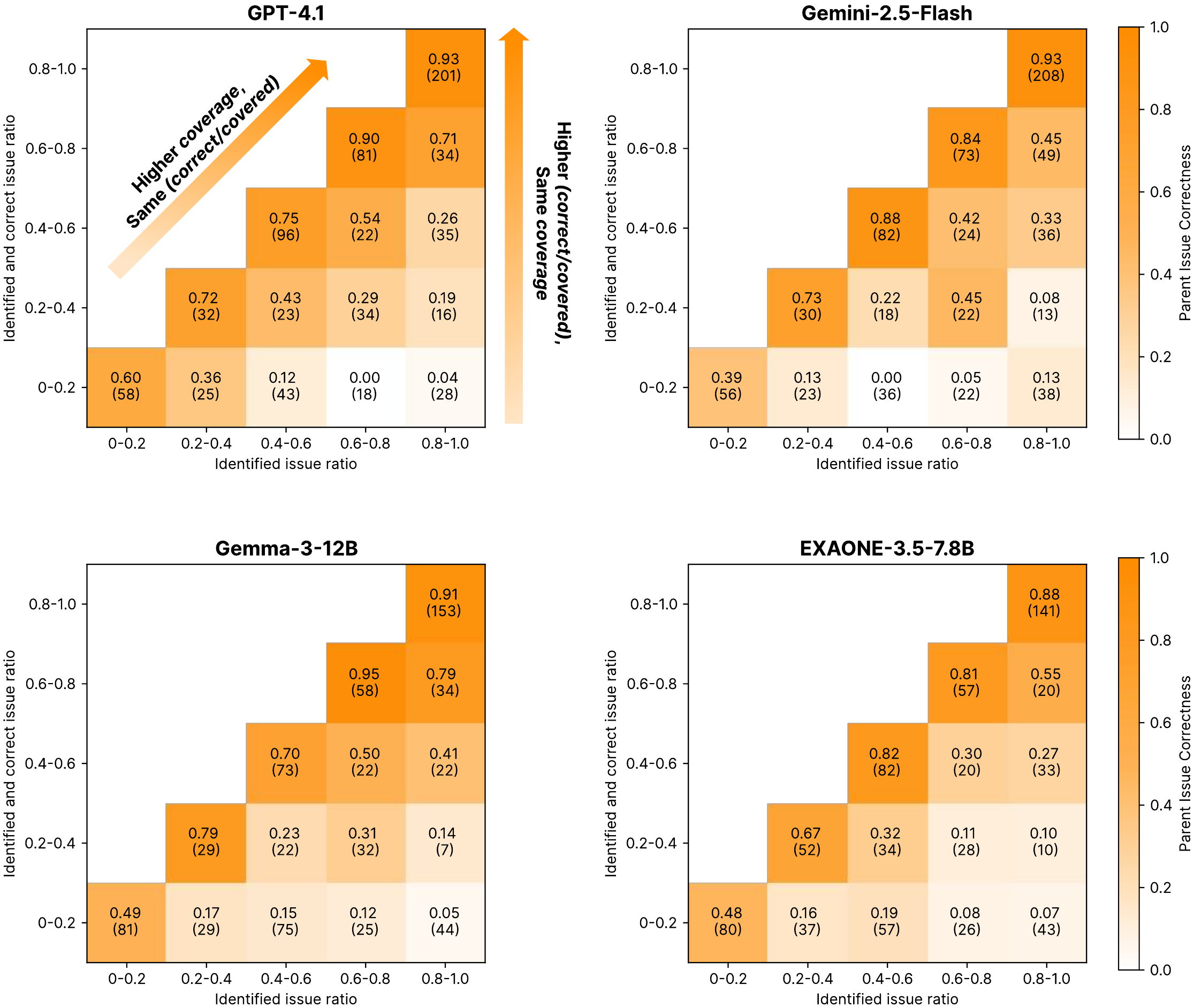}
    \caption{Parent issue correctness binned by child issue coverage and correctness, extending Figure \ref{fig:effects_of_children_node_results}. In each cell, the numbers without parentheses (above) are the parent issue correctness values, and the numbers within parentheses (below) are the number of elements in the bin.}
    \label{fig:effects_of_children_node_results_2}
\end{figure*}

\paragraph{Score per difficulty} LEGIT test data consists of three subsets, \textit{easy, medium, and hard}, divided based on the number of issues. Figure \ref{fig:difficulty} shows how different LLMs perform on each difficulty bin. All models show a consistent drop in LEGIT score, in all three subsections (final order correctness, issue coverage, issue correctness). This indicates that approximating the difficulty of the given case by the number of issues is a plausible strategy.

\paragraph{Issue depth and coverage/correctness} We plot the relation between issue depth and their coverage/correctness in Figure \ref{fig:accuracy_per_depth}. Both coverage and correctness exhibit a clear pattern, where issue coverage decreases rapidly as the depth increases, while issue correctness of covered issues ($\frac{\mathrm{correct}}{\mathrm{covered}}$) decreases as the depth decreases. These observations suggest that these rubrics are inherently structured as a tree, supporting the backward chaining intuition presented in Section \ref{sec:dataset} and motivating analyses in Section \ref{sec:performance}. Once the LLM fails to identify a parent node, it will likely fail to cover its children that must be obtained by decomposing the parent. Similarly, once the child issue is incorrect, the parent will also likely be incorrect, as shown in Figure \ref{fig:effects_of_children_node_results}.

\paragraph{Issue coverage/correctness and final order correctness} Figure \ref{fig:trace_quality_and_final_order_corr} shows the relationship between the reasoning trace's quality (issue coverage and correctness) and the final order correctness. The strong correlation suggests that a high-quality reasoning trace is essential for deriving the correct final order, demonstrating the structural nature of legal reasoning.

\paragraph{Parent issue correctness and child coverage/correctness} We include detailed visualizations about the relationships between parent correctness and children coverage/correctness in Figure \ref{fig:effects_of_children_node_results_2}, extending Section \ref{sec:performance} and Figure \ref{fig:effects_of_children_node_results}. Specifically, we bin all covered parent issues by child coverage and correctness, both ranging from 0 to 1, and evaluate parent issue correctness for each bin. Note that the correctness value is always smaller than coverage, as coverage counts all \textit{covered} issues but correctness counts all \textit{covered and correct} issues. In this figure, the main diagonal represents the contour line of issue correctness for the covered issues ($\frac{\mathrm{correct~child}}{\mathrm{covered~child}}$).

The results show that the parent accuracy decreases when (1) child issue coverage decreases ($\swarrow$) and (2) correctness of covered child issues decreases ($\downarrow$). However, the latter exhibits more severe degradation than the former, consistent with the intuition that low correctness is more detrimental than low coverage (Figure \ref{fig:effects_of_children_node_results}). It is noticeable that the gap between identifying 0-20\% of child issues and 20\%-40\% of child issues is similar to or larger than the gap between 20\%-40\% and 80\%-100\%, motivating the separate definition of decomposition error (identifying no child issues).

\section{Retrieval-augmented generation details}
\label{sec:appendix-rag}

\subsection{Experimental settings}

\paragraph{Retrieval base preprocessing} We use LLMs to extract any citation to statutes and Supreme Court cases, and filter out any malformed strings (the quote being shorter than 20 characters). Then, we apply the MinHash and LSH algorithm \citep{indyk_approximate_1998} implemented by \citet{zhu_ekzhudatasketch_2025} for deduplication. We set the number of permutations in MinHash to 64, and the LSH threshold to 0.65.

\paragraph{BM25 details} The text is first segmented into a list of words, where only content words (nouns, verbs, adjectives, adverbs, numbers, foreign characters) are maintained using the Korean part-of-speech tagger Kiwi \citep{_kiwi_2024}. Then, we use Okapi BM25 \citep{robertson_simple_1994} implemented by \citet{brown_dorianbrownrank_bm25_2025} to retrieve the top 10 candidates using the facts as the query. We use $k_1=1.5$ and $b=0.75$ for hyperparameters.

\paragraph{mContriever details} We use mContriever checkpoint fine-tuned on the multilingual MS-MARCO dataset (\texttt{facebook/mcontriever-msmarco}). We truncate both query and target documents to 512 tokens, which is the maximum length supported by the model.

\paragraph{mContriever fine-tuning details} Extending the results in Section \ref{sec:improving}, we further fine-tune the mContriever checkpoint above using the LEGIT train split. To apply contrastive loss, we use ground-truth citations as positive documents and BM25 retrieval results that are not positive as negative documents. We train the model for 2,000 steps (approx. 3 epochs), with a batch size of 64 and a learning rate of 1e-4.

\subsection{Additional results}

\begin{figure}
    \centering
    \includegraphics[width=\linewidth]{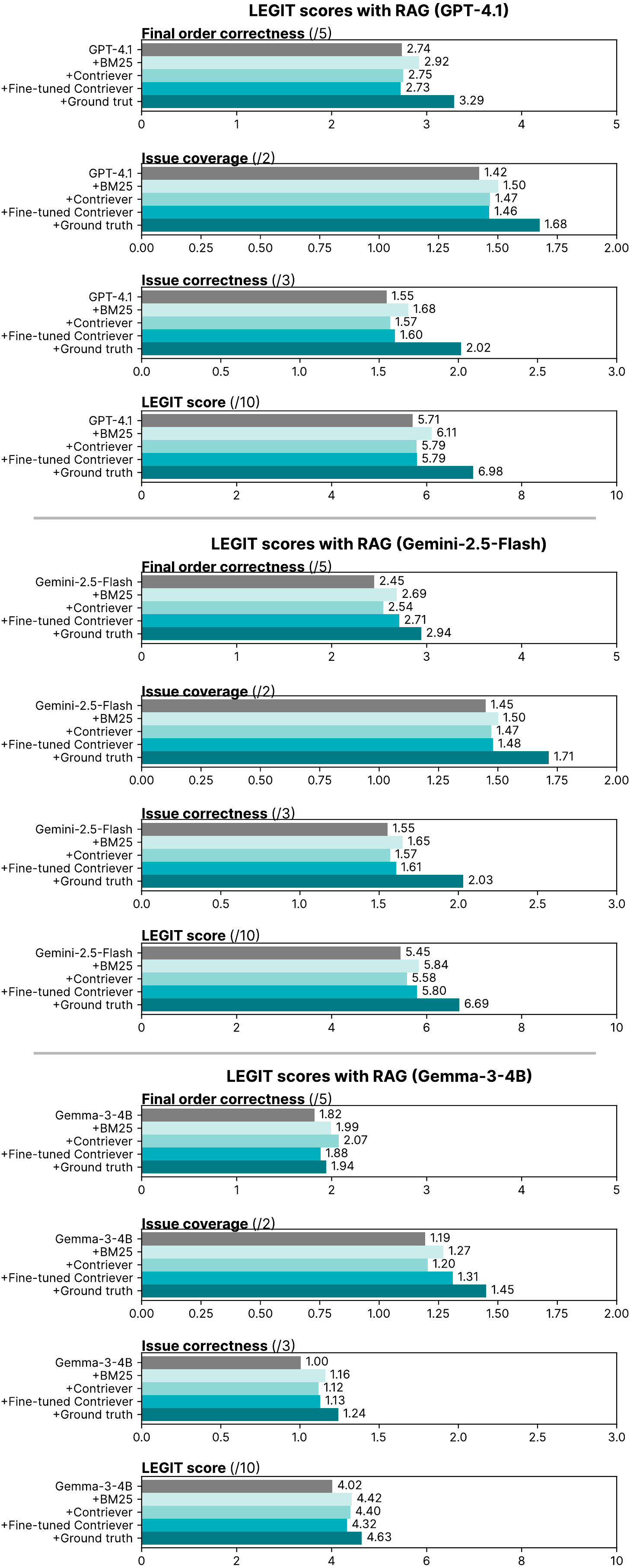}
    \caption{Results of RAG for LEGIT dataset, with three different generators and five retrieval settings (No RAG, BM25, two Contrievers, and Ground-truth citations). RAG improves LEGIT score for around 0.1-0.4 for all (generator, retriever) pairs, with a gain in all three components.}
    \label{fig:legit_score_rag_full}
\end{figure}

\paragraph{RAG improves all three components of LEGIT score.} Figure \ref{fig:legit_score_rag_full} shows the LEGIT score for all three generators (GPT-4.1, Gemini-2.5-Flash, and Gemma-3-4B) and five retrieval settings (No RAG, BM25, Contriever, Fine-tuned contriever, and Ground-truth citations). The LEGIT score increases from 0.1-0.4 for all combinations for three retrievers, and 0.6-1.3 for ground-truth citations. The improvement happens within all three components of the LEGIT score, consistent with the results from Section \ref{sec:improving}.


\paragraph{RAG remains helpful even when the retriever's performance is limited.} Table 1 shows the retrieval performance (Recall@10, NDCG@10) of different retrievers alongside their RAG performance on three generators. Unfortunately, the performance of retrievers is generally low, as directly performing citation retrieval only using the facts is extremely challenging. However, contrary to common belief \citep{yoran_making_2024, amiraz_distracting_2025}, we observe a significant performance gain with RAG in all three generator LLMs despite the low retriever performance. For instance, BM25 achieves a Recall@10 of around 4\%, but the LEGIT scores increase by 0.38-0.42 for all generators, which is between 32.1-64.4\% of the performance gain observed with ground truth citations compared to the base setting. While these retrieved laws might not directly relate to the given situation, we conjecture that prepending relevant laws triggers the \textit{persona effect} \citep{olea_evaluating_2024} that elicits the legal reasoning ability of the LLMs.


\section{Reinforcement learning details}
\label{sec:appendix-rl}

\begin{table}[tb]
    \centering
    \footnotesize
    \begin{tabular}{cc}
        \toprule
        \textbf{Hyperparam.} & \textbf{Value} \\
        \midrule
        Objective & GRPO \\
        KL Div. Coef. & 1e-3 \\
        Max prompt len. & 2048 \\
        Max output len. & 4096 \\
        Batch size & 32 \\
        Rollouts & 8 \\
        Optimizer & AdamW \\
        Learning rate & 1e-6 \\
        Evaluation & every 20 steps \\
        Early stop & 60 steps \\
        \bottomrule
    \end{tabular}
    \caption{Hyperparameters used for RL training, both LEGIT rewards and final order correctness rewards.}
    \label{tab:rl-hyperparams}
\end{table}

\subsection{Experimental settings}

\paragraph{RL settings} For online RL, we use \texttt{verl} \citep{sheng_hybridflow_2025}, an open-source library for training LLMs with RL. We use FSDP as the model training backend, and vLLM for generating rollouts. For reproducibility, we list core training hyperparameters in Table \ref{tab:rl-hyperparams}. For computing training/validation reward with Gemma-3-27B, we use vLLM as the inference engine.

\paragraph{Computation} For training, we use 4 NVIDIA A100 GPUs for model training (rollout generation, backpropagation, \textit{etc.}) and another 4 NVIDIA A100 GPUs for running the online LLM-as-a-judge evaluation. Training with LEGIT reward took approximately 41.6 hours of wall clock time in total, while training with the final order correctness reward took 18.6 hours.

\subsection{LEGIT rewards and final answer rewards}

We compare the behavior of models trained with (1) final order correctness rewards, where the model receives a reward of +10 only when the final order prediction is correct and 0 otherwise, and (2) LEGIT score rewards. Note that both rewards are computed using LLM-as-a-judge. Furthermore, we impose a degeneration penalty of -5 for both settings, where the LLM detects and penalizes undesirable behaviors like code switching and n-gram repetition.

\paragraph{LEGIT score reward achieves better final order accuracy and trace quality than final answer reward.} When training Gemma-3-4B with LEGIT rewards, we observe a significant performance gain in the test set (4.02$\rightarrow$4.77), outperforming RAG with ground truth citations and roughly matching the performance of Gemma-3-27B (4.82). However, training with final answer-only rewards achieves 4.31. Interestingly, the final order correctness of the model is lower than one trained with LEGIT rewards despite being directly trained on that metric. The results show that optimizing the quality of reasoning traces via RL leads to better outcomes in legal reasoning, contrary to the final answer-only RL paradigm in math and programming popularized by DeepSeek-R1 \citep{guo_deepseek-r1_2025}. Furthermore, we show that high inter-LLM consistency (Section \ref{sec:reliability}) transfers to robust RL performance against using different evaluators in train/test time.

\section{Miscellaneous (author checklists)}

\paragraph{Copyright and personal information} Korean court judgments are not bound by any copyrights under law (Copyright Act, Article 7). However, most court judgments are not freely distributed but sold by the Court to cover administrative fees. All judgments used in LEGIT are completely anonymized either by the court or by LBOX. All models and software used in this paper permit academic use.

\paragraph{Randomness} To minimize the randomness, the temperature of the generator and evaluator LLMs was set to 0 in all evaluation experiments. All training experiments (fine-tuning mContriever and RL with rubrics) have been performed once due to high experiment cost and limited compute. 

\section{Prompts}
\label{sec:appendix-prompts}
\begin{figure*}
    \centering
    \includegraphics[width=\linewidth]{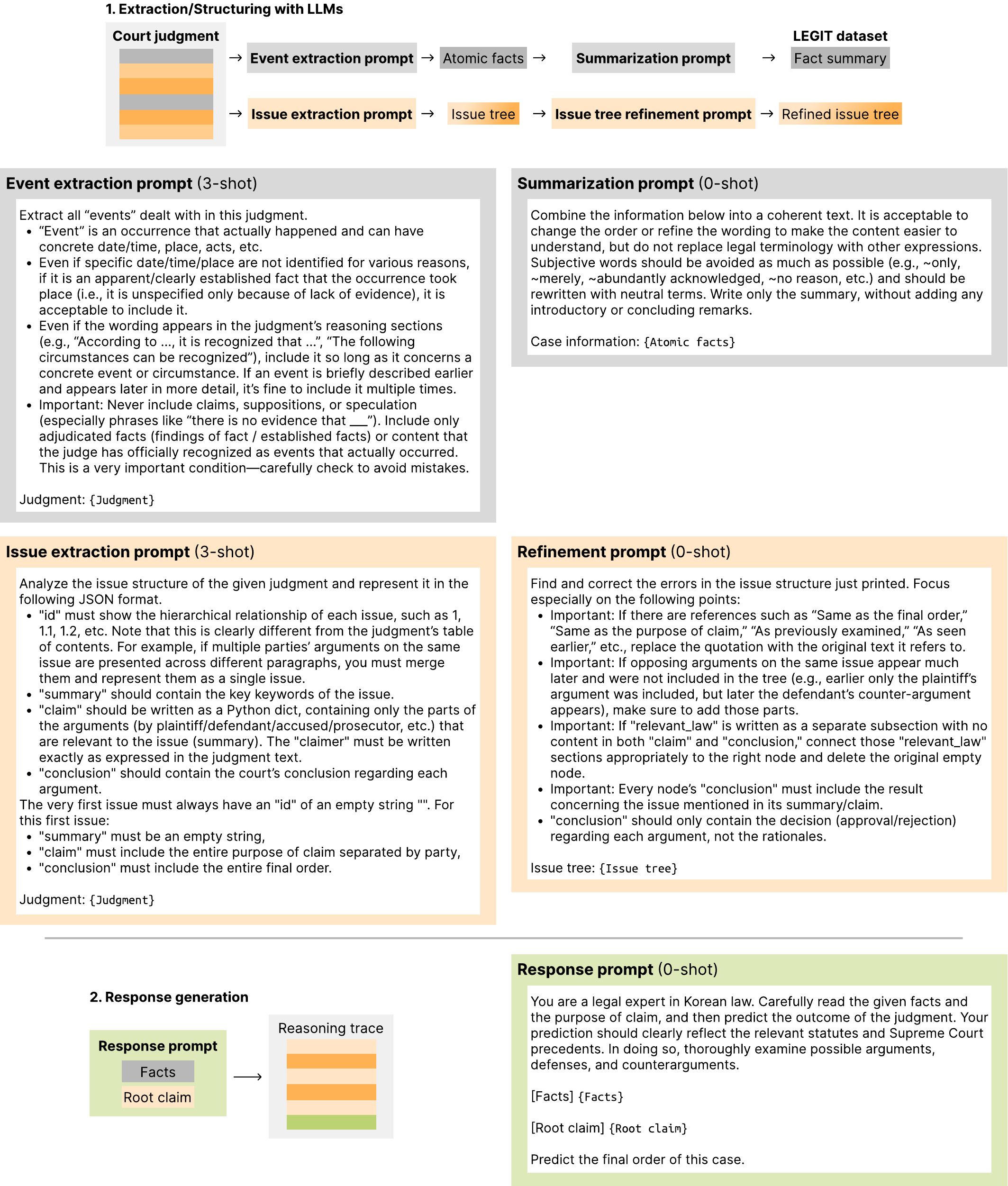}
    \caption{Prompts for dataset construction and generating LLM responses.}
    \label{fig:prompts_1}
\end{figure*}

\begin{figure*}
    \centering
    \includegraphics[width=\linewidth]{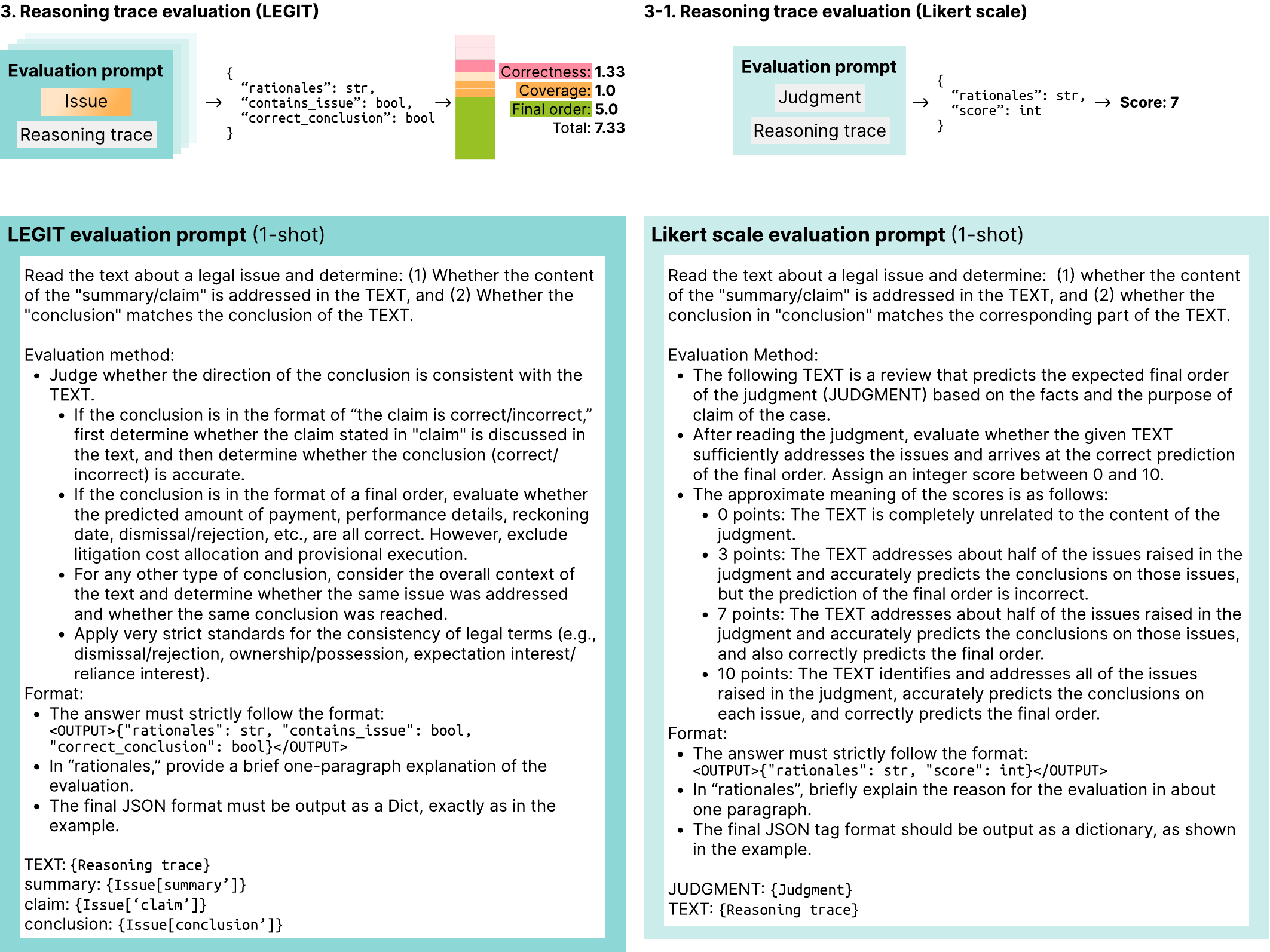}
    \caption{Prompts for evaluating the reasoning traces, either with LEGIT rubrics (left) or Likert scale (right).}
    \label{fig:prompts_2}
\end{figure*}

Figures \ref{fig:prompts_1} and \ref{fig:prompts_2} present the prompts used for creating the LEGIT dataset and evaluating with rubrics. The original Korean prompts are translated into English for readability, and few-shot examples are omitted due to space. The full code and prompts can be found in our official \href{https://github.com/jinulee-v/LEGIT}{GitHub repository}.

\newpage

\newpage

\end{document}